\PassOptionsToPackage{table}{xcolor}
\documentclass[journal,onecolumn]{IEEEtran}

\usepackage{float}
\usepackage[ruled,vlined,linesnumbered]{algorithm2e}
\usepackage{algorithmic}
\usepackage{graphicx,color}
\usepackage{textcomp}
\usepackage{tcolorbox}
\usepackage[caption=false]{subfig}
\usepackage{booktabs}
\usepackage{url}
\usepackage{xcolor}  
\usepackage{amsmath,amssymb,amsfonts}
\IEEEoverridecommandlockouts

\begin{document}

\title{Realistic Urban Traffic Generator using Decentralized Federated Learning for the SUMO simulator}

\author{
    Alberto Bazán-Guillén, Carlos Beis-Penedo, Diego Cajaraville-Aboy, Pablo Barbecho-Bautista, Rebeca P. Díaz-Redondo, Luis J. de la Cruz Llopis, Ana Fernández-Vilas, Mónica Aguilar Igartua and Manuel Fernández-Veiga%
    \thanks{A. Bazán-Guillén, L. de la Cruz Llopis and M. Aguilar Igartua are with the Networking Engineering Department, Universitat Politècnica de Catalunya (UPC), Barcelona, Spain (Emails: alberto.bazan@upc.edu, luis.delacruz@upc.edu, monica.aguilar@upc.edu)}%
    \thanks{D. Cajaraville-Aboy, C. Beis-Penedo, R. Díaz-Redondo, A. Fernández-Vilas, and M. Fernández-Veiga are with the atlanTTic Research Center (I\&C Lab), University of Vigo, Spain (Emails: \{dcajaraville,cbeis,rebeca,avilas,mveiga\}@det.uvigo.es)}%
    \thanks{P. Barbecho-Bautista is with the Department of Electrical, Electronic and Telecommunications Engineering, University of Cuenca, Cuenca, Ecuador (Email: pablo.barbecho@ucuenca.edu.ec)}%
    
    \thanks{Corresponding author: Alberto Bazán-Guillén.}%
}

\maketitle

\begin{abstract}
Realistic urban traffic simulation is essential for sustainable urban planning and the development of intelligent transportation systems. However, generating high-fidelity, time-varying traffic profiles that accurately reflect real-world conditions, especially in large-scale scenarios, remains a major challenge. Existing methods often suffer from limitations in accuracy, scalability, or raise privacy concerns due to centralized data processing.
This work introduces DesRUTGe (Decentralized Realistic Urban Traffic Generator), a novel framework that integrates Deep Reinforcement Learning (DRL) agents with the SUMO simulator to generate realistic 24-hour traffic patterns. A key innovation of DesRUTGe is its use of Decentralized Federated Learning (DFL), wherein each traffic detector and its corresponding urban zone function as an independent learning node. These nodes train local DRL models using minimal historical data and collaboratively refine their performance by exchanging model parameters with selected peers (e.g., geographically adjacent zones), without requiring a central coordinator.
Evaluated using real-world data from the city of Barcelona, DesRUTGe outperforms standard SUMO-based tools such as RouteSampler, as well as other centralized learning approaches, by delivering more accurate and privacy-preserving traffic pattern generation.
\end{abstract}


\section{INTRODUCTION}
\label{sec:introduction}

\IEEEPARstart{T}{he} escalating challenges of urban mobility, intensified by global sustainability goals and the evolution of smart city paradigms, increase the need of advanced tools for traffic management and planning. Realistic urban traffic simulation is paramount for developing and validating intelligent transportation systems (ITS), optimizing public transport, assessing the impact of new mobility services, and redesigning urban spaces. Open-source  simulators such as SUMO (Simulation of Urban MObility) \cite{SUMO} offer granular control over individual vehicle dynamics and enable the modeling of complex and city-scale traffic scenarios. However, the generation of traffic patterns that realistically capture temporal and spatial variability in urban environments remains a considerable challenge within current simulation frameworks. Although SUMO provides mechanisms for traffic definition \cite{behrisch2011sumo}, automatically achieving high fidelity with observed traffic dynamics, especially across diverse urban zones and throughout a full 24-hour cycle, is a complex problem that often requires substantial manual effort or relies on simplified statistical models \cite{barbecho2021stgt}.

Addressing this challenge requires learning complex spatio-temporal traffic behaviors from available data, often collected from distributed urban sensing infrastructure such as inductive loop detectors, magnetometers, cameras with computer vision, radar, and Bluetooth or Wi-Fi tracking systems. Traditional centralized machine learning approaches, where all data is gathered and processed on a single server, face significant obstacles in this domain. These include potential privacy violations due to the sensitive nature of mobility data \cite{kairouz2021advances}, high communication overhead from transmitting large amounts of raw sensor readings \cite{li2020federated}, and scalability issues as the number of data sources or the size of the urban area grows. 

The application of Deep Reinforcement Learning (DRL) has shown considerable promise for learning complex control policies in dynamic environments, including traffic signal optimization \cite{wei2019presslight}. DRL agents can learn to make sequential decisions to achieve specific objectives by interacting with an environment, such as a traffic simulator \cite{arulkumaran2017deep}. However, training DRL agents to model city-wide phenomena typically requires extensive data or coordinated multi-agent learning strategies, which, when centralized, raise significant concerns regarding privacy and scalability.

On the other hand, Federated Learning (FL) has emerged as a compelling alternative, enabling collaborative model training across multiple distributed clients (e.g., traffic sensors or computational nodes representing city zones) without sharing their raw local data, inherently enhancing data privacy and reducing communication costs \cite{mcmahan2017communicationefficient}. However, standard FL often still relies on a central orchestrator, which can introduce single points of failure, hindering robustness. Decentralized Federated Learning (DFL) takes this a step further by eliminating the central server entirely, so the clients communicate directly with a subset of their peers (e.g., geographical neighbors) to exchange model information and perform aggregation locally or collectively. This serverless architecture offers enhanced fault tolerance, can improve scalability by localizing communication, and further reduces reliance on a central entity, aligning well with the distributed nature of urban sensor networks \cite{savazzi2020federated}.

To address the limitations of existing traffic generation methods, this paper proposes a methodology for the automatic generation of realistic urban traffic patterns using only minimal statistical information, such as the average hourly traffic intensity over a 24-hour period. In this work, we introduce DesRUTGe (Decentralized Realistic Urban Traffic Generator), a novel SUMO-based tool that employs decentralized DRL to model and reproduce realistic, time-varying vehicle densities that closely align with real-world traffic profiles obtained from monitoring stations. DesRUTGe adopts a DFL architecture in which individual monitoring stations independently train local models and collaborate by sharing knowledge, thus improving collective learning while ensuring scalability and privacy. Our approach demonstrates higher accuracy and faster convergence compared to existing SUMO-compatible methods for generating 24-hour variable traffic profiles.

The main contributions of this work are the following.
\begin{itemize}
    \item \textbf{Decentralized and Privacy-Preserving Framework for Traffic Generation:} We propose \textit{DesRUTGe}, a novel DFL framework for generating realistic urban traffic profiles using DRL agents within SUMO. Unlike most FL or Federated Reinforcement Learning (FRL) applications in Intelligent Transportation Systems (ITS), which typically focus on prediction or control tasks and rely on centralized aggregation, DesRUTGe adopts a fully decentralized architecture based solely on communication between neighboring nodes. This design preserves data privacy by sharing only model parameters locally and reduces communication overhead while still enabling effective collaborative learning across regions.
    \item \textbf{Closed-Loop Traffic Generation with SUMO and Open Data:} We leverage SUMO as a closed-loop environment for DRL-based traffic generation. By injecting vehicle routes and capturing feedback through SUMO’s virtual detectors, the agent learns to modulate traffic flows to match specified target intensities. The entire pipeline relies exclusively on open data sources, such as OpenStreetMap \cite{OSM} and historical traffic counts, making it fully adaptable to diverse urban scenarios without the need for proprietary software or infrastructure.
    \item \textbf{Localized DRL for Adaptive Traffic Synthesis:} Each DRL agent is trained to synthesize traffic patterns that reflect local real-world conditions within its assigned urban zone, thereby facilitating adaptation to the city’s heterogeneous traffic dynamics.
    \item \textbf{Empirical Validation and Performance:} We demonstrate through simulations using real-world data from Barcelona (Spain) that DesRUTGe can generate 24-hour traffic patterns with high fidelity, significantly outperforming existing SUMO tools, particularly during peak congestion periods.
\end{itemize}

The remainder of this paper is organized as follows. Section~\ref{sec:background} provides background on SUMO, Reinforcement Learning, and Decentralized Federated Learning. Section~\ref{sec:related_work} reviews related literature. Section~\ref{sec:methodology} details the proposed DesRUTGe methodology. Section~\ref{sec:experimental setup} describes the experimental setup, followed by results and discussion in Section~\ref{sec:results}. Finally, Section~\ref{sec:conclusions} concludes the paper and outlines future work.

\section{BACKGROUND}
\label{sec:background}

The design of smart mobility services in urban environments requires both accurate traffic modeling and intelligent decision-making mechanisms. On the one hand, traffic simulators such as SUMO provide a flexible and highly detailed platform for replicating real-world vehicular behaviors at microscopic scales. On the other hand, recent advances in Machine Learning (ML), Reinforcement Learning (RL), and FL allow for decentralized, data-driven decision-making under uncertainty and privacy constraints. This section highlights the convergence of these technologies as a basis for developing intelligent, privacy-conscious services in emerging urban mobility systems.

\subsection{Simulation of Urban MObility}
\label{subsec:sumo}

In the context of vehicular communication simulations, accurately representing vehicle mobility within the environment is essential to produce meaningful and reliable results. A prominent tool in this domain is the Simulation of Urban MObility (SUMO) \cite{SUMO}, an open-source, highly portable, microscopic traffic simulation platform developed by the German Aerospace Center (DLR). As an open-source project, SUMO provides a flexible and extensible framework that allows researchers and urban planners to model, test, and analyze a wide variety of mobility scenarios in complex urban environments.

Microscopic traffic simulators like SUMO are designed to simulate the movement of individual vehicles, capturing their interactions and dynamics with a high degree of precision. This fine-grained modeling is particularly valuable for applications such as ITS, autonomous vehicle development, and studies focused on electric vehicle behavior, including State of Charge (SoC) analysis. Unlike macroscopic simulators that aggregate traffic flows, SUMO's microscopic approach enables a more realistic representation of behaviors such as lane-changing, acceleration, deceleration, and driver decision-making \cite{krajzewicz2012recent}.

To model vehicle mobility, SUMO implements several models, most notably a variation of the Krauss car-following model \cite{kraussBib}, which has been extended to support multi-lane behavior, lane-changing decisions, and collision avoidance mechanisms. This flexibility allows researchers to tune vehicle dynamics for specific use cases or integrate more sophisticated behavior models as needed.

Moreover, SUMO includes a variety of tools for generating traffic demand, either through random trips, predefined traffic flows, or OD (origin-destination) matrices. Vehicles can be instantiated with detailed attributes by using device containers that store data on emissions, fuel/energy consumption, battery usage, and communication capabilities. This modular structure is particularly beneficial for evaluating sustainable urban mobility strategies, such as the deployment of the electric vehicle fleet or traffic decarbonization policies \cite{SUMO}.

The open nature of SUMO, combined with its active development community and wide adoption in both academia and industry, makes it a key enabler for urban mobility research. It does not only facilitate reproducibility and transparency in simulation studies but also integrates smoothly with other open-source tools, such as network simulators (e.g., OMNeT++ \cite{omnet}), geographic data systems (e.g., OpenStreetMap), and machine learning frameworks.

\subsection{Reinforcement Learning}
\label{subsec:rl}

Within the ML spectrum, RL is a powerful framework for sequential decision-making. In RL, agents learn optimal behaviors by interacting with an environment to maximize cumulative rewards over time~\cite{barto2021reinforcement}. Unlike supervised learning, RL does not rely on labeled datasets but instead uses trial-and-error strategies, guided by policies that map observed states to actions.

Formally, any RL problem can be modeled as a Markov Decision Process (MDP), defined by a tuple $(\mathcal{S}, \mathcal{A}, \mathcal{P}, R, \gamma)$, where:
\begin{itemize}
    \item $\mathcal{S}$ is the set of possible states representing the environment's configuration at a given time.
    \item $\mathcal{A}$ is the set of available actions the agent can take.
    \item $\mathcal{P}$ is the state transition probability function, which defines the dynamics of the environment.
    \item $R$ is the reward function, which provides scalar feedback to guide the learning process.
    \item $\gamma \in [0,1]$ is the discount factor, which determines the importance of future rewards.
\end{itemize}

In partially observable environments, agents receive only partial information about the true state, known as an observation. In these cases, policies must be learned from incomplete data, increasing the complexity of the decision-making process. The agent’s goal is to learn a policy $\pi(a|s)$ that maximizes the expected long-term return $R_t$, see Eq. (\ref{eq:reward}) \cite{puterman2014markov}
\begin{equation}
\label{eq:reward}
    R_t = \sum_{k=0}^{\infty} \left[ \gamma^k \cdot r_{t+k+1} \right],
\end{equation}
where $R_t$ represents the total expected cumulative reward starting at time $t$; $r_{t+k+1}$ is the immediate reward received in the future step $t+k+1$; $\gamma$ is the discount factor that reduces the importance of rewards in the future; the sum adds all future rewards, weighting them by $\gamma^k$, so closer rewards matter more than distant ones.

RL has achieved remarkable successes in areas such as robotics, autonomous systems, and network optimization. However, its practical deployment faces key challenges: (i) sample inefficiency due to the need for extensive exploration; (ii) scalability limitations, especially in large or continuous state–action spaces; and (iii) simulation-to-reality gaps, which arise when transferring models trained in virtual environments to real-world applications.

To address these challenges, DRL has emerged as a powerful approach that combines RL algorithms with deep neural networks. DRL enables agents to approximate value functions or policies in high-dimensional, unstructured or continuous state–action spaces where traditional tabular methods are infeasible and classical RL suffers from scalability and sample-efficiency limitations~\cite{mnih2015human}. Techniques such as Deep Q-Networks (DQN), Proximal Policy Optimization (PPO) and Actor-Critic models have significantly expanded the applicability of RL to complex domains, including urban mobility, smart grids and multi-agent systems.

\subsection{(Decentralized) Federated Learning}
\label{subsec:federated_learning}

In recent years, FL has emerged as an alternative strategy to the traditional machine learning paradigm where all training information resides on a single central server~\cite{mcmahan2017communicationefficient}. Driven by the rise of distributed environments, such as Internet of Things (IoT) systems or Edge/Fog computing environments, the centralized FL (CFL) scheme orchestrates a collaborative and iterative learning process where each participant/client trains a local ML model using only its own local data. After the local training phase, the updated models (or their gradients) are transferred to a server, which aggregates the parameters using schemes such as Federated Averaging (FedAvg) \cite{FedAvg}. Finally, the parameter server redistributes the aggregated version for the next round of training.

As an alternative to CFL, the promising DFL paradigm~\cite{beltran2023decentralized} has recently emerged, where the role of the central server is eliminated. Instead, model aggregation is performed among the clients themselves, who exchange their parameters with a subset of peers (e.g., close neighbors or neighbors with related statistical characteristics), generating different communication topologies. This scheme enhances fault tolerance as there is no single point of failure and scales more naturally as the number of participants grows. However, decentralization entails a higher cost in model communication~\cite{liu2022decentralized}, a challenge that is addressed by lightweight dissemination protocols (e.g. Gossip protocol) or gradient compression.

Although both approaches avoid direct exposure of each client's local data (decreasing data communication costs), FL presents different vulnerabilities studied in depth in the literature~\cite{bouacida2021vulnerabilities,xie2024survey}. In the privacy field, global model/gradient inversion threats and membership inference attacks stand out, often addressed by differential privacy or secure multiparty computing techniques~\cite{lyu2022privacy}. In the security field, the diversity of devices and distribution of the learning process allows the possibility of adversarial behavior by malicious nodes. These are known as Byzantine attacks and different state-of-the-art statistical techniques exist to filter or mitigate these attacks~\cite{shi2022challenges}. DFL intensifies these security challenges by not having a central coordination point, increasing the surface of Byzantine attacks. To address these challenges, there are different techniques such as specific Byzantine-robust methods or blockchain techniques~\cite{fang2024byzantine, li2020blockchain}, among others. 

In FL environments, the most commonly trained learning algorithms are neural networks (NN), due to their many applications in fields such as computer vision or natural language processing~\cite{zhang2021survey}. NN are easily parameterized, allowing the application of statistical aggregation algorithms based on these model parameters. Extending this method to other ML algorithms, such as decision trees, is non-trivial and requires more complex aggregation methods~\cite{heiyanthuduwage2024decision}. Recent research has explored the integration of RL in FL environments, resulting in a new paradigm known as Federated Reinforcement Learning (FRL)~\cite{zhu2021federated}. FRL allows distributed agents to learn policies locally by interacting with their individual environments, while periodically sharing abstract knowledge (e.g., gradients, Q-values, or policy parameters) with a central server or among peers. This scheme allows aggregation of these contributions to obtain improved policies, preserving privacy and supporting on-device learning for privacy-sensitive domains.

\section{RELATED WORK}
\label{sec:related_work}

Recent years have witnessed significant advances in traffic modeling and intelligent mobility systems, driven by the integration of high-fidelity simulators, machine learning, and distributed intelligence. This section reviews relevant literature on three main fronts: the use of traffic simulation tools in smart city environments, the application of Reinforcement and Deep Reinforcement Learning to traffic optimization problems, and the emerging role of Federated Learning as a privacy-preserving paradigm in distributed traffic data generation and decision-making.

\subsection{Traffic Simulation Tools in Smart Cities}
\label{subsec:toolSim}

Accurate traffic simulation is essential to evaluate intelligent transportation systems, urban mobility planning, and the deployment of autonomous and connected vehicles in smart city environments. Several simulation frameworks have emerged to address this need, varying in their modeling granularity, scalability, and integration capabilities.

Simulation tools like SUMMIT \cite{cai2020summit} exemplify how urban mobility models can integrate road infrastructure and dynamic traffic behaviors specific to densely populated environments. These tools have been successfully used to assess autonomous driving systems in complex urban settings, particularly where vehicle-pedestrian interactions are highly unpredictable. Similarly, SceneGen \cite{tan2021scenegen} employs auto-regressive neural networks to generate realistic traffic scenarios without relying on rigid rule-based systems, offering a scalable approach to simulate behaviorally rich urban environments for autonomous vehicles.

Among the most well-known traffic simulation tools, Simulation of Urban MObility (SUMO) \cite{SUMO} is widely recognized as a leading open-source microscopic traffic simulator, supporting highly detailed modeling of vehicle movements, interactions, and traffic infrastructure. SUMO allows for various modes of traffic generation, including random trip generation, fixed route assignments, dynamic traffic assignment (DTA), and reproducible origin-destination (OD) matrices derived from real-world traffic data \cite{behrisch2011sumo}. These capabilities have made SUMO particularly useful in traffic congestion analysis \cite{smith2014sumo}, eco-route planning \cite{sagaama2024energy}, and infrastructure planning.

What makes SUMO stand out is its integration flexibility. Its Traffic Control Interface (TraCI) enables dynamic modification of simulation parameters and vehicle routes during run-time, making it ideal for the study of adaptive and large-scale systems. Additionally, SUMO tools such as \texttt{randomTrips.py}, \texttt{OD2TRIPS}, and \texttt{DUAROUTER} facilitate the creation of both stochastic and deterministic traffic scenarios, providing high versatility for experiments requiring varied or repeated conditions.

In a similar vein, the traffic model proposed in \cite{RealisticTrafficValencia} leverages real-time sensor data from induction loop detectors to generate high-fidelity Origin-Destination (OD) matrices, yielding more precise route distributions and flow estimations. This method outperforms SUMO’s built-in \texttt{dfrouter} tool, offering improved realism in urban-scale simulations.

Moreover, SUMO's open architecture and compatibility with other urban simulation platforms, such as OMNeT++ \cite{omnet}, Veins \cite{veins}, or MATLAB \cite{matlab}, have enabled researchers to integrate mobility dynamics with communication and energy systems, fostering cross-domain experimentation in smart city contexts. Compared to proprietary or commercial simulators (for example, VISSIM \cite{vissim} or Aimsun \cite{aimsun}), SUMO offers unparalleled flexibility, cost-effectiveness and reproducibility, key aspects in scientific research and real-world deployment scenarios.

\subsection{Traffic Modeling through Reinforcement Learning}
\label{subsec:RLtrafficGeneration}

RL and DRL have become prominent tools in traffic modeling, particularly for solving sequential decision-making problems in dynamic and uncertain environments such as urban road networks. Their ability to learn adaptive strategies from interaction with simulated or real traffic environments makes them well-suited for traffic control, route planning, and resource allocation.

One of the most studied applications of RL in traffic systems is traffic signal control. Early approaches applied tabular Q-learning or SARSA algorithms to individual intersections \cite{arel2010reinforcement}, while more recent works utilize Deep Q-Networks (DQNs) or actor-critic frameworks for large-scale coordination of multiple intersections \cite{wei2019presslight, van2016coordinated}. These models have shown superior performance over traditional fixed-time or rule-based methods, especially in reducing average waiting times and improving throughput.

The use of real-world data has also emerged as a critical component for realistic traffic simulation. For instance, in \cite{arias2017prediction}, a Markov-chain-based traffic model was employed to forecast electric vehicle (EV) charging demand, incorporating real-time traffic data sourced from CCTV cameras in Seoul. This work demonstrates how integrating spatiotemporal traffic dynamics contributes to context-aware simulation environments and supports sustainable infrastructure planning.

Human behavior modeling has further enriched traffic simulation research. The approach in \cite{cao2024reinforcement} combines reinforcement learning with human feedback to train agents that simulate not only realistic driver behavior but also adherence to traffic laws. This human-centered reinforcement learning highlights the necessity of embedding behavioral realism in simulation frameworks to reflect real-world complexities more accurately.

Another growing application area is platooning and cooperative driving, where multi-agent DRL (MARL) techniques are employed to manage vehicle coordination in dense traffic scenarios, improving safety and traffic flow \cite{isele2018navigating}. Similarly, DRL has been employed for controlling complex driving behaviors such as lane-changing and ramp merging under dense traffic conditions. For instance, Shalev-Shwartz et al. \cite{shalev2016safe} propose a multi-agent DRL framework that ensures safety in highway scenarios.

Beyond traffic signal and routing control, high-fidelity simulators such as CARLA \cite{dosovitskiy2017carla} have enabled the use of DRL in complex urban driving tasks. CARLA provides a 3D photorealistic environment equipped with various sensors and supports dynamic interactions with vehicles, pedestrians, and environmental conditions. It has been instrumental in training end-to-end DRL agents for urban navigation, collision avoidance, and behavioral cloning.

Several recent studies have focused on the use of reinforcement learning for traffic simulation and control. Proximal Policy Optimization (PPO) in particular has shown promising results in tasks involving dynamic, non-linear environments such as urban traffic networks. 

In a previous study by our team \cite{guillenrutge}, we introduced a preliminary approach in which a DRL-based agent was trained using Proximal Policy Optimization (PPO) to generate realistic traffic flows across the city of Barcelona. This initial system relied on aggregated traffic intensity data, averaging measurements from all detectors, and employed a single global agent to manage traffic across the entire network. Although this approach demonstrated the feasibility of using reinforcement learning for realistic traffic generation, it lacked spatial specificity and failed to capture local variations in traffic behavior across different urban zones. Building upon that methodology, the present work introduces a decentralized training framework. Rather than relying on a single global agent, each traffic detector is assigned to a region—defined via Voronoi partitioning—and associated with an independent DRL agent. This decentralized setup enables fine-grained control and a more accurate reproduction of local traffic conditions. Additionally, inter-agent model exchange strategies are proposed and analyzed to enhance training robustness and generalization capabilities.

Despite promising results, RL-based traffic modeling still faces challenges related to generalization, sample inefficiency, and transferability to real-world deployments. Research continues on hierarchical RL, curriculum learning, and sim-to-real adaptation to bridge this gap. In this context, we adopt SUMO due to its open-source nature, high customizability, and support for fine-grained traffic modeling based on real detector data. Unlike proprietary traffic simulators, SUMO offers full control over route generation, detector integration, and scenario configuration, making it particularly well-suited for training and validating RL agents in a transparent and reproducible environment.

\subsection{Federated Learning Applications in Traffic Generation}
\label{subsec:fl_traffic_gen}

The integration of FL offers a promising avenue for leveraging distributed data sources in Intelligent Transportation Systems (ITS) while upholding data privacy. Initial applications predominantly focused on traffic prediction tasks, demonstrating early works the viability of FL for traffic flow forecasting by sharing model parameters instead of raw data, achieving comparable accuracy to centralized methods \cite{Liu2020FLTraffic}. Subsequent research explored enhancing these FL systems by incorporating blockchain for decentralized aggregation to remove single points of failure \cite{Qi2021BlockchainFL} or by developing personalized FL approaches to tackle data heterogeneity across different urban zones or cities \cite{Zhang2024CrossCityFL}. While these studies underscore the benefits of collaborative, privacy-preserving learning for analytical tasks in transportation, they typically rely on a central server for model aggregation or address decentralization through mechanisms different from direct, localized peer-to-peer exchanges.

More recently, the concept of truly DFL, which obviates the need for any central coordinator, has begun to permeate ITS research. A notable example is the work by Shen et al., who proposed a DFL-based spatial-temporal model for freight traffic speed forecasting where regional clients exchange model updates exclusively with their geographical neighbors \cite{Shen2024DecentralizedFL}. This approach aligns with the architectural principles of DesRUTGe, demonstrating the feasibility of neighbor-based model fusion for traffic-related tasks. However, the application of such DFL mechanisms has largely remained within the domain of predictive modeling.

The task of generating synthetic data, particularly comprehensive traffic scenarios for simulation, using federated paradigms is an even less explored frontier. Some research has ventured into using FL to train generative models for data augmentation purposes, for example, by collaboratively training a diffusion model to create synthetic traffic trajectories to enhance the performance of traffic flow prediction models \cite{Orozco2025FedTPS}. While this represents a step towards ``federated synthesis'' \cite{little2023federated}, the primary goal remains data augmentation for predictive tasks, and these systems often still employ a centralized FL architecture. Other traffic scenario generation tools, while aiming for realism and diversity, typically rely on centralized data collection and model training, without incorporating distributed or federated learning principles \cite{feng2023trafficgen}.

In parallel, Federated Reinforcement Learning (FRL) has shown promising results in traffic control applications, particularly in optimizing traffic signals across multiple intersections. These approaches allow RL agents at different intersections to share knowledge (e.g., model gradients or parameters) to learn coordinated control policies, often outperforming isolated RL agents \cite{ye2021fedlight}. While these FRL systems involve distributed RL agents, they predominantly utilize a central server for aggregation, or hierarchical structures with cluster-level coordinators \cite{fu2025federated}, rather than direct neighbor-to-neighbor DFL. Furthermore, their objective is to generate optimal control policies, not the underlying traffic demand patterns.

To the best of our knowledge, the specific challenge of generating realistic 24-hour urban traffic profiles for detailed simulation—using DRL agents within a truly decentralized, neighbor-only DFL framework—remains largely unexplored. Our proposed system, DesRUTGe, addresses this critical gap by combining the adaptive learning capabilities of DRL agents operating in localized geographic zones with a decentralized federated learning (DFL) scheme, in which agents share and aggregate model updates exclusively with their immediate neighbors. This approach aims to leverage the benefits of collaborative learning to generate high-fidelity, city-wide traffic patterns, while enhancing privacy, scalability, and robustness by eliminating any central point of control or data aggregation. Methodological advancements in generic DFL, such as proxy model sharing for serverless learning \cite{kalra2023decentralized}, further support the timeliness and feasibility of such fully decentralized architectures. Our work, therefore, distinguishes itself by applying these advanced DFL principles to the challenging generative task of realistic urban traffic synthesis through distributed DRL. Table~\ref{tab:related_work_comparison} summarizes the reviewed works, highlighting their primary tasks, learning paradigms, federation types, and key distinctions with respect to our proposed approach, DesRUTGe.

\begin{table*}[ht]
\centering
\caption{Comparison of Selected Related Works with DesRUTGe}
\label{tab:related_work_comparison}
\resizebox{\textwidth}{!}{%
\begin{tabular}{|l|l|l|l|l|l|}
\hline
\textbf{Reference} & \textbf{Primary Task} & \textbf{Learning Paradigm} & \textbf{Federation Type} & \textbf{Aggregation} & \textbf{Key Difference from DesRUTGe} \\ \hline
\cite{Liu2020FLTraffic} & Traffic Prediction & ML (GRU) & FL & Centralized & Task (Prediction); Centralized \\ \hline
\cite{Shen2024DecentralizedFL} & Traffic Prediction & ML (S-T Model) & DFL & Neighbor-Only & Task (Prediction); Not DRL-based generation \\ \hline
\cite{Orozco2025FedTPS} & Data Augmentation & Generative (Diffusion) & FL & Likely Centralized & Goal (Augmentation); Not DRL; Likely Centralized \\ \hline
\cite{feng2023trafficgen} & Traffic Generation & Generative (Autoregressive) & None & N/A & No FL/DFL; Centralized data/model \\ \hline
\cite{ye2021fedlight} & Traffic Control & DRL (A2C) & FRL & Centralized & Task (Control); Centralized Aggregation \\ \hline
\cite{fu2025federated} & Traffic Control & DRL & Hierarchical FRL & Cluster-level + Global & Task (Control); Not peer-to-peer DFL \\ \hline
\textbf{DesRUTGe (Ours)} & \textbf{Traffic Generation} & \textbf{DRL (PPO)} & \textbf{DFL} & \textbf{Neighbor-Only} & \textbf{Focus on generation with DRL in true DFL} \\ \hline
\end{tabular}%
}
\end{table*}

\section{METHODOLOGY}
\label{sec:methodology}

Building upon the framework introduced in \cite{guillenrutge}, the proposed approach, DesRUTGe, retains the general mechanism of DRL-based traffic generation using SUMO and PPO. However, it significantly extends the system’s flexibility and realism by adopting a decentralized learning strategy and incorporating detector-specific targets derived from real-world data. For completeness and clarity, we provide here a comprehensive technical description of the whole methodology.

The primary objective of this work is to develop a realistic traffic generation framework—intended for use with simulators such as SUMO—that is capable of reproducing historical traffic intensity patterns observed within a given urban region. By leveraging reinforcement learning techniques alongside real-world traffic data, the proposed methodology aims to generate vehicular flows that closely replicate the temporal and spatial characteristics of actual traffic measurements collected within an urban environment. The following sections describe the simulation environment, data processing steps, traffic generation pipelines (over one hour and one day), the reinforcement learning strategies employed to achieve this goal, and the decentralized approach in order to ensure scalability and privacy.

\subsection{SUMO simulation for traffic generation}
\label{subsec:callSUMO}

The generation of baseline traffic in our simulations is centered on controlling the number of moving vehicles in designated urban zones for each target time interval. This process requires two essential components: a mechanism to inject vehicles into the network and a feedback system to measure resulting traffic densities. Together, they enable a closed-loop control framework to achieve desired traffic conditions across space and time.

The traffic generation procedure was developed using the SUMO simulator and draws inspiration from the STG tool \cite{barbecho2021stgt} developed by our research team. The goal is to simulate traffic for a given hour in a specific city area by injecting a predefined number of vehicles into the road network. These vehicles follow routes between randomly selected origin-destination (OD) pairs to create traffic flows that match the desired intensity levels. Traffic in SUMO is defined by a route file (\texttt{.rou.xml}), which specifies for each vehicle its unique identifier, departure time and position, and the sequence of road segments (edges) it traverses to reach its destination. To ensure proper traffic distribution and avoid routing conflicts, a valid set of origins, destinations, and departure times must be carefully generated.

Prior to route generation, the simulation area must be delineated. In SUMO, this is typically done using Traffic Analysis Zones (TAZs), which are predefined areas represented as sets of edges (street segments) within the map. Traffic levels are measured using SUMO’s built-in detectors, among which the \textit{induction loop}  is one of the most commonly used. These devices emulate real-world magnetic loop detectors and can capture data such as vehicle count, traffic intensity, and occupancy at specific locations, usually placed near intersections.

The traffic generation pipeline begins with the creation of OD matrices, specifying the number of vehicles to be generated between pairs of TAZ-defined zones. OD pairs are randomly and uniformly distributed within the selected area to meet the desired traffic volume. These OD matrices are then converted into trips using SUMO’s \texttt{od2trips} tool. In SUMO, a \textit{trip} is defined as a simplified description of a vehicle's journey, containing only the origin edge, destination edge, and departure time, without intermediate routing details.

Subsequently, the trips are transformed into complete routes using SUMO’s \texttt{duarouter}, which performs shortest-path calculations to determine the edge sequences vehicles will follow. The resulting routes are then included in a simulation configuration file \texttt{(.sumocfg)}, which also specifies the network file (\texttt{.net.xml}), detector configuration \texttt{(.add.xml)}, and output files for traffic data collection.

Once the simulation is executed, the output files defined in the \texttt{.add.xml} configuration are analyzed to extract data from the \textit{induction loops}. These files report hourly vehicle counts and traffic intensity levels at each measurement point. The recorded intensity values serve as the reference for the traffic control loop, ensuring that subsequent iterations can adjust traffic generation parameters to maintain targeted congestion levels.

\subsection{Reinforcement Learning Algorithm}
\label{subsec:RLtraining}

Accurately estimating the number of vehicles to inject at any given time in a traffic simulation, in order to achieve a desired traffic intensity, is a non-trivial task. Urban traffic dynamics are inherently chaotic, and the topology of city networks—characterized by complex road layouts and intersection behaviors—renders analytical or rule-based approaches infeasible. Moreover, the limited availability of labeled traffic data makes supervised learning techniques unsuitable. In contrast, RL presents an attractive alternative, as it does not require labeled data and instead leverages interactions between an agent and an environment to learn optimal behaviors over time.

\vspace{0.5cm}

\subsubsection{Reinforcement Learning Definition}
\label{sec:RL_definition}

In this context, we propose a model-free RL framework in which an agent iteratively adjusts the number of vehicles injected into a SUMO simulation to match a predefined target traffic intensity. This target is measured at specific locations within the city using \textit{induction loop} detectors. The primary objective is to minimize the error between the observed traffic intensity and the target profile, which represents the desired traffic conditions for a specific urban region and time window.

\vspace{0.5cm}

The model-free RL framework consists of the following components:

\begin{itemize} \item \textbf{State ($\mathcal{S}$)}: The current estimate of the number of vehicles to inject into the map is denoted by state $s$, which belongs to the state space $S$. This is a natural number and does not necessarily correspond to the actual number of vehicles detected in the simulation. To enhance the efficiency and stability of the reinforcement learning process, the state $s$ was normalized to the range $[0, 1]$ using the minimum and maximum observed values for each traffic metric. This normalization ensures consistent scaling and prevents any single feature from dominating the learning process.

\item \textbf{Action (\(\mathcal{A}\))}: A scalar adjustment factor applied multiplicatively to the current state. Actions are sampled from a continuous range \(a \in [-0.1, +0.3]\), enabling the agent to incrementally increase or decrease the vehicle count according to Eq. (\ref{eq:stateaction})
\begin{equation}
\label{eq:stateaction}
    s' = s \cdot (1 + a),
\end{equation}
where \(s\) denotes the current number of vehicles, \(s'\) is the new state and \(a \in \mathcal{A}\) is the action taken.

\item \textbf{Environment}: The SUMO simulator, which processes the injected traffic and returns traffic measurements obtained from the \textit{induction loops} sensors.

\item \textbf{Observation (\(O\))}: The average traffic intensity reported by \textit{induction loops} after running the simulation with a given \(s'\). This traffic intensity is compared against the target traffic intensity to evaluate the performance of the new state \(s'\).

\item \textbf{Reward (\(R\))}: A scalar signal derived from the absolute error between the observed traffic intensity (\(O\)) and the predefined target traffic intensity (\(T\)). This formulation encourages the agent to minimize the deviation from the desired traffic state:
\begin{equation}
\label{eq:Reward}
R = - \left| T-O \right| 
\end{equation}
\end{itemize}

To steer the agent towards convergence and promote efficient solutions, we enhance the design of the reward signal by incorporating two additional components, whose specific values were tuned after extensive testing:

\begin{itemize}
\item \textbf{Step Penalty (\(\lambda\)):} A value of \(\lambda = 0.01\) was chosen to apply a mild penalty at each step, encouraging the agent to pursue efficient solutions without overshadowing the primary objective.

\item \textbf{Goal Achievement Reward (\(\eta\)):} A large positive reward of \(\eta = 10.0\) was granted when the agent achieved a mean squared error smaller than \(threshold = 0.001\). This strong positive signal prioritized policies that met the desired precision.
\end{itemize}

Finally, the designed reward function is formally defined as follows:
\begin{equation}
R_{\text{total}} =
\begin{cases}
\label{eq:RewardTotal}
    R - \lambda, & \text{if } \left| T-O \right| \geq 10^{-3} \\
    +\eta, & \text{if } \left| T-O \right| < 10^{-3}
\end{cases}
\end{equation}

The use of SUMO as the RL environment allows for realistic feedback on the effects of actions. Since SUMO realistically simulates traffic dynamics given a set of initial conditions, it serves as a reliable model for evaluating the agent’s policy decisions. The system is entirely model-free and does not require an explicit transition probability function, as the dynamics are implicitly learned through interaction with the realistic environment.

This setup enables the agent to learn a policy $\pi(s)\to a$ that gradually converges to injecting a number of vehicles, resulting in traffic densities that closely match the target traffic profile, thus achieving fine-grained calibration of urban mobility simulations in data-scarce environments.

\subsubsection{Traffic Calibration through Proximal Policy Optimization}
\label{sec:PPO}

Proximal Policy Optimization (PPO) \cite{schulman2017proximal} was selected as the reinforcement learning algorithm for this work due to its robustness, sample efficiency, and compatibility with environments like SUMO, where agent-environment interactions are computationally expensive. PPO uses a clipped surrogate objective to ensure stable policy updates and reliable learning without the need for complex optimization techniques. Its ability to generalize well and operate effectively under high-dimensional conditions makes it particularly suitable for our traffic generation task. In this context, PPO is used to train a policy that determines how many vehicles to inject into the simulation to closely match predefined traffic targets, learning from episodic simulations of one hour in specific urban regions.
\vspace{0.5cm}

In a nutshell, the following steps are carried out to calibrate the traffic profile:

\begin{enumerate}
  \item During training, the agent observes the current state $s$ using outputs from SUMO.
  \item The agent samples an action $a$ applying the policy, adjusting the number of vehicles and transitioning to a new state $s'$ as defined in~\eqref{eq:stateaction}.
  \item SUMO is executed with this updated configuration and the resulting vehicle counts obtained from the \textit{induction loops} are retrieved.
  \item A reward $R_{\text{total}}$ is then computed according to~\eqref{eq:RewardTotal} based on the observed values and the target traffic intensity.
  \item Finally, the PPO algorithm updates the policy using cumulative rewards to minimize deviation from the target traffic.
\end{enumerate}

\vspace{0.5cm}

The PPO-based reinforcement learning agent is trained over multiple episodes using a variety of traffic intensity targets sampled from real-world data covering a typical 24-hour weekday. A goal-conditioned reward structure—penalizing deviations from the target and rewarding accuracy—guides the learning process, enabling the agent to efficiently converge toward the desired traffic profile. Through this training regime, the agent acquires a robust and generalizable policy capable of generating realistic traffic patterns for one-hour intervals that closely approximate the expected traffic densities across diverse urban scenarios.

\subsection{Traffic Generation algorithms using the trained model}

Once the PPO agent is trained to accurately generate realistic traffic patterns for one-hour intervals, it can be employed as a callable component for calibrating traffic over longer time periods. The user supplies a vector of hourly target values representing the desired traffic densities across a full 24-hour period.

\vspace{0.5cm}

\subsubsection{Traffic Generation for a Single Hour}

To generate the traffic pattern for a specific hour $h$, the agent is invoked using the procedure referred to as the Traffic Generation Algorithm (TGA), which is outlined in Algorithm~\ref{alg:algo1h}.

\begin{algorithm}[tb]
\caption{1h-TGA $\Rightarrow$ 1-hour Traffic Generation Algorithm.}
\label{alg:algo1h}
\SetAlgoLined 
\KwData{Hour $h$; target intensity \texttt{target[$h$]}.}
\KwResult{Route file (\texttt{routes.rou.xml}); traffic residual vector (\texttt{residual[]}).}
Initialize \texttt{routes.rou.xml} as an empty file\;
Initialize \texttt{residual[]} as a vector of zeros\;
\(threshold = 0.001\)\; 
\While{$\lvert \textnormal{observed traffic intensity} - target[$h$] \rvert > \texttt{threshold}$}
{
    Call PPO model with \texttt{target[$h$]}, state $S$ to propose action $a$ and adjust traffic injection\;
    Update observed traffic intensity and calculate reward $R(s,a)$\;
    Make $s \rightarrow s'$\ using~\eqref{eq:stateaction}; 
}
Compute \texttt{residual[]} as the amount of traffic delayed to subsequent hours\;
\Return \texttt{routes.rou.xml}, \texttt{residual[]}\;
\end{algorithm}

The process begins with the initialization of the simulation environment and an empty route file (lines 1 and 2), which will ultimately store the vehicle routes required to reproduce the target traffic for hour $h$. The initial state $s_0$ represents an initial estimate of the number of vehicles to be injected. In each iteration of the control loop, SUMO is executed with the current state $s$, and the resulting traffic densities are observed through \textit{induction loop} detectors deployed at strategic locations throughout the simulation scenario.

The agent then compares the current observation to the hourly target (line 4). If the absolute difference between the observed traffic intensity and the target value exceeds a predefined threshold of $10^{-3}$ (set in line 3), the PPO agent is queried for a new action $a \in \mathcal{A}$, which is applied to update the state to $s'$ using \eqref{eq:stateaction}. This updated vehicle injection value is then used in the next simulation iteration. The loop continues until the observation meets the precision criteria, i.e., the absolute difference between the observed traffic intensity and the target value falls below the accepted threshold.

In addition to generating the main observation for the hour of interest, the simulation also outputs residual traffic effects extending into subsequent hours (line 9). This occurs because some vehicle routes span multiple hours due to network congestion or long trips. To account for this phenomenon, the simulation returns a residual traffic vector of length 23, representing the traffic densities observed in the 23 hours following hour $h$. This vector is critical for chaining multiple hourly simulations, as it ensures continuity and consistency of traffic flow across the entire 24-hour window.

At the end of the process, the system returns two outputs (see lines 9 and 10): (i) the completed SUMO route file (\texttt{*.rou.xml}), which contains the vehicle trips reproducing the desired traffic pattern for hour $h$; and (ii) the residual traffic vector, which is passed on to subsequent hourly simulations.

\vspace{0.5cm}

\subsubsection{24-Hour Traffic Generation}

Once the PPO-based traffic generation model is prepared to accurately reproduce traffic over a one-hour window, it can be employed to generate a full 24-hour traffic profile. This is accomplished by sequentially invoking the one-hour traffic generator (see Algorithm~\ref{alg:algo1h}) for each hour of the day, while accounting for residual traffic effects carried over from previous hours. The procedure is outlined in Algorithm~\ref{alg:algo24h}.

\begin{algorithm}[hb]
\caption{24h-TGA $\Rightarrow$ 24-hours Traffic Generation Algorithm.}
\label{alg:algo24h}
\SetAlgoLined 
\KwData{Map file (\texttt{.net.xml}); start time $T_{\text{start}}$; end time $T_{\text{end}}$; hourly target intensities in the specific city (\texttt{target[]}).}
\KwResult{Complete route file (\texttt{final\_routes.rou.xml})}.
Initialize \texttt{final\_routes.rou.xml} as an empty file\;
\For{$i \gets T_{\text{start}}$ \KwTo $T_{\text{end}}$}{
    \tcp{Generate traffic for hour $i$}
    \texttt{[routes.rou.xml, residual[]] = 1h-TGA($i$, target[$i$]) $\leftarrow$ Alg.~\ref{alg:algo1h}}\; 
    Append \texttt{routes.rou.xml} to \texttt{final\_routes.rou.xml}\;
    \tcp{Adjust targets for subsequent hours}
    \For{$j \gets i+1$ \KwTo $T_{\text{end}}$}{
        \texttt{target[$j$] = target[$j$] - residual[$j$]}\;
    }
}
\Return \texttt{final\_routes.rou.xml}\;
\end{algorithm}

For each hour \( h \in \{0, 1, \ldots, 23\} \), the corresponding target traffic vector—derived from measured traffic data provided by the Barcelona City Council in our case study—is used as input to the trained reinforcement learning agent. The agent then generates vehicle injection patterns aimed at reproducing the target traffic intensity as accurately as possible at the monitored intersections equipped with \textit{induction loops}.

After each simulation, the output provides not only the observed traffic for the current hour but also a residual traffic vector extending over the subsequent 23 hours. This residual component reflects vehicles that remain in the system due to prolonged travel times or congestion, thereby influencing traffic intensities beyond the hour in which they were injected.

To ensure consistency and realism across consecutive simulations, the residual traffic is incorporated into the target adjustment process. Specifically, the traffic volume carried over from the previous hour is subtracted from the original target of the current hour. This adjustment prevents over-injection of vehicles and reflects a more accurate vehicle flow across time steps. The updated target is computed using Eq. (\ref{eq:adjusted-target}):
\begin{equation}
\label{eq:adjusted-target}
\text{AdjustedTarget}_{h} = \text{Target}_{h} - \text{Residual}_{h},
\end{equation}
where \(\text{Residual}_{h}\) denotes the portion of traffic from previous injections that is still active during the hour \(h\).

This iterative approach ensures that each hourly simulation incorporates the dynamic effects of previous traffic injections, resulting in a coherent 24-hour traffic evolution. By sequentially linking consecutive one-hour simulations, a comprehensive set of vehicle routes—spanning different departure times and destinations—is generated. The resulting route file captures the spatiotemporal dynamics of urban mobility, providing a realistic and data-driven representation of traffic flow throughout the day.

This methodology effectively balances precision of local traffic reproduction with global temporal consistency, enabling accurate assessments of mobility interventions and infrastructure planning under realistic daily traffic patterns.

\subsection{Federated Approach for Traffic Generation}
\label{subsec:framework}

\begin{figure*}[tph]
    \centering
    \includegraphics[width=0.8\linewidth]{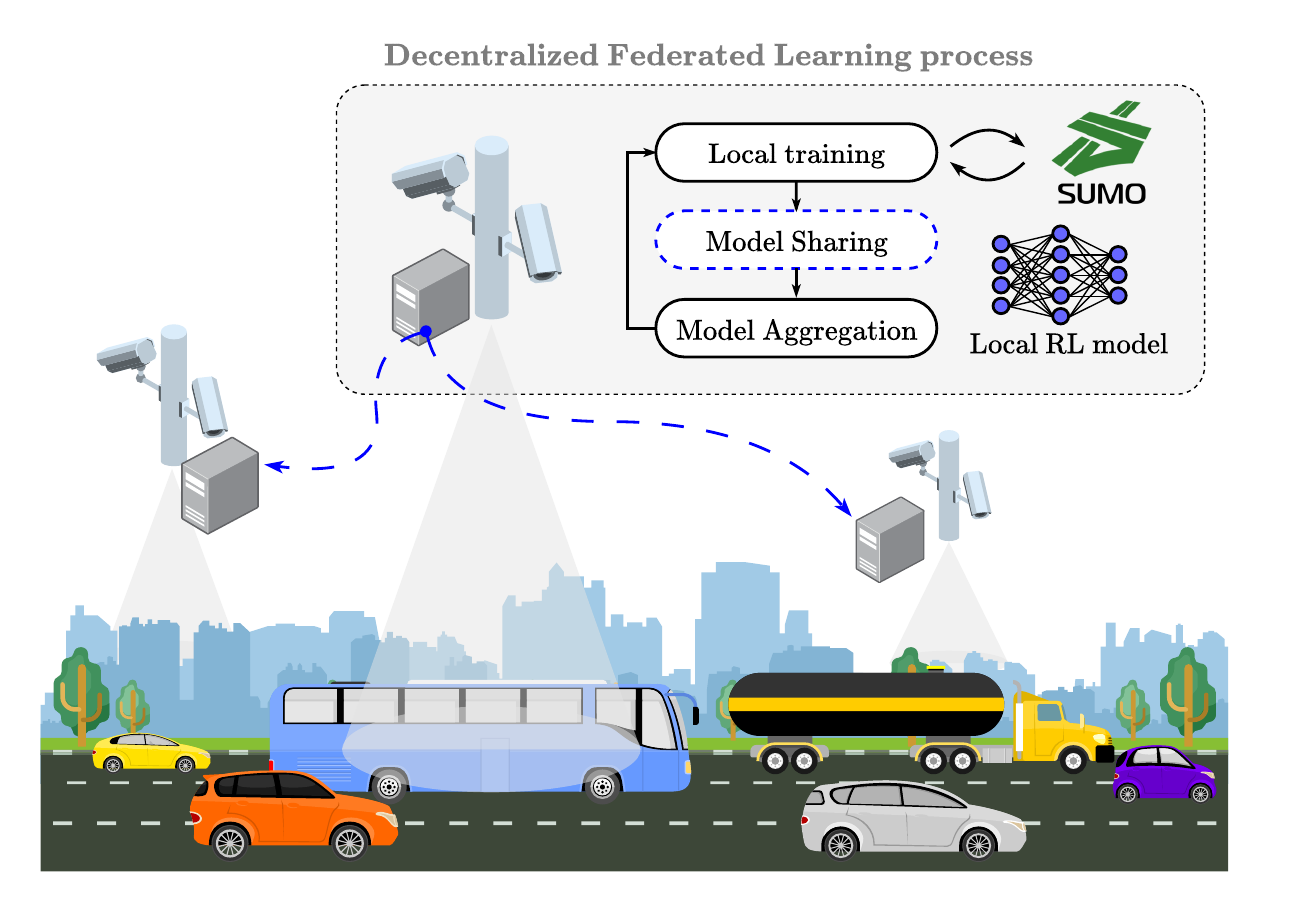}
    \caption{Decentralized Federated Learning Scenario for Urban Traffic Generation in the SUMO Simulator.}
    \label{fig:methodology}
\end{figure*}

After presenting the necessary modifications to the SUMO simulator and the characteristics of the RL learning algorithm in the centralized scenario, this subsection introduces DesRUTGe, our proposed decentralized learning framework for synthetic traffic generation in smart cities. DesRUTGe allows multiple detectors within the same city to perform collaborative training of the previously described RL model. This approach: (i) improves data privacy because local information is not shared or centralized; (ii) increases the system's scalability and robustness because it does not depend on a central server; and (iii) enhances more generalizable and robust learned policies by aggregating experience across agents operating in different urban zones. Figure~\ref{fig:methodology} shows the proposed framework and scenario, where the different agents, in charge of obtaining data on traffic intensity in different areas of a city, carry out a collaborative learning process.

We model this distributed learning scenario using a DFL approach, in which each traffic detector trains its own learning model with its local data, and then exchanges its local RL model with corresponding neighbors (i.e., an FRL scenario). The topology of a static DFL scheme can be modeled as an undirected graph $G=\left( \mathcal{V},\mathcal{E} \right)$, where $\mathcal{V}$ represents the city's detectors as vertices, and $\mathcal{E} \subseteq \mathcal{V} \times \mathcal{V}$ represents the neighborhood relationships between the detectors. The criteria used to define neighborhoods between detectors can vary and are not limited to geographic proximity. For instance, agents may be grouped based on similar traffic intensities or clustered according to shared traffic patterns.

The DFL framework consists of three fundamental phases executed during each federation round: (i) local training, (ii) model sharing, and (iii) aggregation process. This iterative process takes place over multiple rounds, with the exact number depending on the application context and defined objectives. The following paragraphs define each of these phases in detail as applied to our scenario.

\vspace{0.5cm}

\textbf{[1] Local training:} Each detector in the city measures traffic intensity and stores the corresponding data. To ensure independent coverage, the city is partitioned into non-overlapping regions, with each region assigned to a specific traffic detector. Using the traffic intensity data collected within its designated zone, a RL model is trained locally, as described in Subsection~\ref{subsec:RLtraining}. This setup enables each model to effectively learn the specific traffic patterns and characteristics of its assigned zone. During each federation round, every agent trains its local RL model by executing multiple instances of the SUMO simulator, involving several simulation steps.

\textbf{[2] Model sharing:} Once the traffic detectors finish their local training, they share their local model with their neighbors in the topology according to the established criteria. This process enables each detector to learn about the traffic characteristics of the other detectors without sharing any private data, thereby reducing bandwidth consumption and improving privacy. In our learning algorithm, the model parameters correspond to the weights of the neurons in the PPO neural network associated with the RL model. 

\textbf{[3] Aggregation process:} Each traffic detector aggregates the models received from its neighboring detectors, along with its own local model, to produce a single updated model that is subsequently used for training on local data in the next federation round. Given that the PPO parameters are real-valued, various statistical aggregation methods can be employed depending on the characteristics of the trained RL models, such as trimmed mean or coordinate-wise median. In this work, we prioritize system functionality and therefore adopt the classical FedAvg scheme \cite{FedAvg}, which performs a coordinate-wise average of the PPO parameters.

\section{EXPERIMENTAL SETUP}
\label{sec:experimental setup}

In this work, we focus on modeling realistic urban traffic patterns in the city of Barcelona, a metropolitan area known for its dense and heterogeneous traffic dynamics. According to recent studies, Barcelona experiences some of the highest levels of road congestion in Europe during peak hours, with an average travel time of 33 min per 10 km and congestion levels reaching 45\% in certain districts \cite{tomtom2021traffic}. These characteristics make it an ideal case study for evaluating intelligent traffic generation and calibration systems.

\subsection{Map Processing and Ground Truth Definition}

The urban layout of Barcelona was obtained from OpenStreetMap (OSM) \cite{OSM}, an open-source geographic data platform. To convert the map into a format suitable with the SUMO traffic simulator, we used the \texttt{netconvert} tool provided by the SUMO suite \cite{krajzewicz2012recent}. This utility transforms raw OSM files into SUMO-readable road network files with the extension \texttt{.net.xml}, encoding the full set of road segments (edges), intersections (nodes), and their respective attributes. The resulting map reflects the real-world road geometry and topology, enabling high-fidelity simulations.

\begin{figure}[tb] 
\centering 
\includegraphics[width=0.8\linewidth]{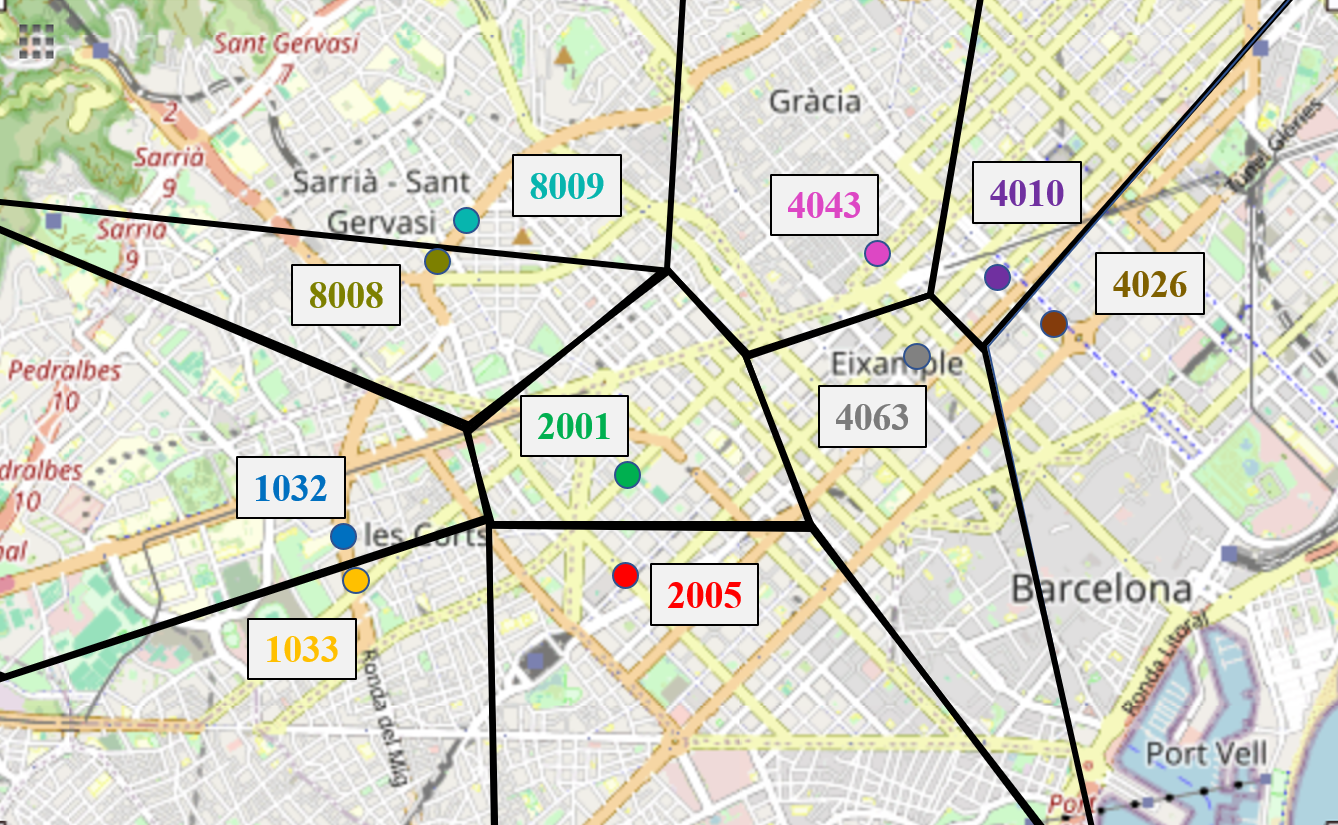} \caption{Simulation scenario of a 45 km$^2$ area in Barcelona (9 km x 5 km) showing the locations of ten traffic detectors and a Voronoi-Based TAZ Division. Map sourced from OSM \cite{OSM}.} 
\label{fig:voronoi_taz} 
\end{figure}

To define the target traffic densities for each simulation zone, we used real-world vehicle count data provided by the Barcelona City Council. The dataset contains hourly vehicle counts collected from ten strategically located detectors distributed throughout the city, whose identifiers are: 1032, 1033, 2001, 2005, 4010, 4026, 4043, 4063, 8008, and 8009. These traffic detectors are permanently installed at key intersections and continuously monitor traffic volumes using magnetic or infrared sensors. An illustrative view of their locations is shown in Fig~\ref{fig:voronoi_taz}.

Based on this dataset, we derived weekday traffic profiles from 2019—the most recent period unaffected by pandemic-related disruptions. For each detector, we computed the mean hourly vehicle count across all weekdays, yielding a 24-dimensional target vector per site. These vectors constitute the reference traffic patterns that the RL agent aims to replicate via optimized vehicle injection strategies.

\subsection{SUMO integration}

To enable traffic observation within SUMO, virtual induction loops were deployed at the same locations as the real-world detectors. These simulated sensors emulate vehicle detection hardware, allowing SUMO to record vehicle counts and other traffic metrics at specific coordinates in the simulation map. Each loop captures the number of vehicles passing through its location during every simulated hour, thus providing a direct analog to the physical sensors and enabling quantitative performance comparisons.

To spatially structure the traffic generation process, the city map was partitioned into distinct zones corresponding to the area of influence of each detector. Each zone, or TAZ, encompasses the road segments that are most geographically and topologically relevant to its associated detector.

To ensure fair and spatially coherent partitioning, we applied a Voronoi tessellation based on the geographic coordinates of the detectors. A Voronoi diagram ensures that the perimeter of each Voronoi polygon is equidistant from its neighbors, creating non-overlapping regions where each zone contains exactly one detector and all nearby streets. This method is particularly suitable for our use case as it respects spatial proximity and avoids arbitrary administrative boundaries. Figure~\ref{fig:voronoi_taz} shows the resulting partitioning of the Barcelona map into ten TAZs.

This structure enables localized traffic generation and calibration: for each TAZ, the reinforcement learning agent operates independently, generating vehicle flows that match the corresponding 24-hour target profile. This spatial decomposition reduces the overall complexity of the task and facilitates the scalable training and deployment of decentralized agents.

In real-world urban environments, traffic data is typically collected from multiple strategically placed detectors. To accurately replicate realistic traffic conditions across a city, it is essential to generate localized traffic patterns that align with each detector’s historical observations, while maintaining global consistency throughout the entire simulation area.

To achieve this, the entire simulation map is divided into independent subregions, each associated with a single traffic detector. These subregions are both mutually exclusive and collectively exhaustive, meaning each part of the city is assigned to exactly one detector and the entire area is covered. The most suitable method for this spatial division is the use of Voronoi diagrams~\cite{aurenhammer1991voronoi}.

Voronoi diagrams partition a space based on proximity to a set of given points. In our context, each point represents a traffic detector, and the diagram assigns every location in the map to the nearest detector, forming a polygonal region around each one. These regions define the area of influence for each detector and serve as the spatial basis for independent traffic generation processes.
\vspace{0.5cm}

This method offers several advantages:
\begin{itemize}
    \item Each detector governs traffic generation in its local region, preserving spatial relevance.
    \item The full map is covered without overlap, avoiding interference between detectors.
    \item The structure naturally supports modular and parallel processing for large-scale traffic simulations.
\end{itemize}

By employing Voronoi-based partitioning, the proposed approach ensures that traffic generation remains locally accurate while preserving global consistency with real-world data across the urban network.

\subsection{Implementation Details}
\label{sec:implementation}
Our objective is to perform distributed computational simulations using a DFL approach. After reviewing several alternatives in the literature~\cite{Beltran23,kholod2020open}, we selected the open-source simulator DecentralizedFedSim\footnote{Code is available online: \\ \textsf{\url{https://gitlab.com/compromise3/decentralizedfedsim}}}, previously developed in~\cite{cajaraville2024byzantine} by some of the authors of this work. The DecentralizedFedSim simulator allows one to define various distributed learning scenarios, such as non-collaborative, centralized FL, and DFL, thanks to its modular architecture. As a result, users only need to focus on specifying the dataset, the implemented RL model, and the model-sharing topology among nodes. The version of DecentralizedFedSim adapted for this work is publicly available on GitLab\footnote{Code is available online: \\ \textsf{\url{https://gitlab.com/compromise3/desrutge }}}, including all the modifications required to implement our proposed approach.

The modifications made to the simulator include the development of several modules and functions for data preprocessing, the implementation of the RL model, and its training in conjunction with the SUMO traffic simulator. To this end, each agent autonomously issues terminal-level commands to SUMO and associated tools to extract the environmental data required for model adaptation and PPO-based policy optimization.

In addition, other more specific changes to the DFL simulator were required. Specifically, the simulator used multithreading for the management of the different nodes participating in the DFL scenario, which generated conflicts due to multiple SUMO calls in a single instance. For this, the simulation was adapted with the \texttt{multiprocessing} library so that each node is dedicated on a different CPU core. Besides, other modifications were necessary for memory management due to the change from threads to processes.

On the other hand, focusing on the performed simulations, the hyperparameters listed in Table~\ref{tab:ppo_hyperparams} were selected based on a combination of default values recommended in literature and empirical tuning adapted to the traffic generation problem.

\begin{table}[thp]
\centering
\caption{PPO Hyperparameters}
\label{tab:ppo_hyperparams}
\begin{tabular}{lc}
\toprule
\textbf{Parameter} & \textbf{Value} \\
\midrule
Learning rate ($\alpha$) & 0.0003 \\
Steps per update ($n_{\text{steps}}$) & 64 \\
Batch size & 64 \\
Epochs per update & 4 \\
Discount factor ($\gamma$) & 0.95 \\
GAE $\lambda$ & 0.95 \\
Clip range & 0.2 \\
Entropy coef. & 0.01 \\
Value loss coef. & 0.5 \\
Max grad. norm & 0.5 \\
\bottomrule
\end{tabular}
\end{table}

The hyperparameters used for the Proximal Policy Optimization (PPO) algorithm (Table~\ref{tab:ppo_hyperparams}) were selected based on established guidelines from the literature \cite{schulman2017proximal, stable-baselines3}, as well as empirical adaptation to the traffic generation task. A learning rate of 0.0003 is widely adopted in reinforcement learning applications due to its balance between convergence speed and stability. The number of steps per update and batch size were both set to 64, consistent with configurations known to perform well in discrete action spaces with moderately complex environments. The number of epochs per policy update (4) and clipping range (0.2) follow the original PPO recommendations \cite{schulman2017proximal}, which prevent excessive policy shifts and stabilize training. The discount factor $\gamma=0.95$ and GAE $\lambda=0.95$ provide a trade-off between bias and variance in the advantage estimation, as discussed in \cite{schulman2015high}. The entropy coefficient (0.01) encourages exploration, while the value loss coefficient (0.5) and gradient clipping (0.5) contribute to stable critic training and prevent exploding gradients. This configuration, while grounded in standard practice, was also fine-tuned through pilot simulations to adapt to the specific dynamics of vehicle generation in SUMO. 

All simulations and training procedures were conducted on a high-performance Dell PowerEdge R7525 server. The machine is equipped with two AMD EPYC 7713 processors (128 cores, 256 threads in total, 2.0 GHz), 768 GB of DDR4 RAM (12 $\times$ 64 GB at 3200 MT/s), and two 8 TB 7.2K RPM SATA hard drives. An NVIDIA A100 GPU with 80 GB of memory was used to accelerate computation during reinforcement learning.

\subsection{Validation Scenarios}
\label{subsec:decentralized}

To address the complexity of traffic generation in large urban environments, we propose a decentralized learning framework in which each traffic detector and its associated TAZ act as an independent learning node. Each node is responsible for generating realistic hourly traffic in its region using a locally trained PPO agent, following the method detailed in subsection~\ref{subsec:RLtraining}. The training process at each node consists of multiple \textit{rounds}, where each round comprises 100 local training episodes. At the end of each round, nodes perform a model exchange procedure with a selected peer, followed by local aggregation and further training.

In order to preserve decentralization and avoid full synchronization across all nodes (which would reduce the approach to centralized training with higher bandwidth consumption), model exchanges are restricted to selected peers. We evaluate three model-sharing strategies:

\begin{itemize}
    \item \textbf{Geographic Neighbors:} Each node shares its model with neighboring nodes, as defined by a Voronoi diagram constructed from the locations of traffic detectors. Specifically, two nodes are considered neighbors if their corresponding TAZ regions share at least one boundary. This neighborhood definition leverages spatial locality, promoting knowledge transfer between physically adjacent zones that are likely to exhibit similar traffic dynamics.
    
    \item \textbf{Affinity-Based Neighbors (Volume Similarity):} Nodes are grouped based on the magnitude of their historical traffic volumes. Hierarchical clustering using Ward's linkage method~\cite{WardLinkage} with Euclidean distance was applied directly to the raw 24-hour traffic count profiles of each traffic detector. Detectors with similar average daily vehicle counts are clustered together, as illustrated by the dendrogram\footnote{A dendrogram is a type of tree-shaped diagram commonly used to represent the hierarchy of clusters in hierarchical clustering algorithms (such as agglomerative or divisive methods).} in Fig.~\ref{fig:dendro_volume}, and model exchanges occur within each cluster. The underlying hypothesis is that nodes managing similar traffic loads can benefit from sharing strategies to handle traffic intensity. We selected 3 different groups of influence based on the results of the dendogram at the cut point of 2000 vehicles/hour, to have 3 clusters with different traffic intensities: 

    \begin{itemize}
    \item Group of traffic detectors with Low traffic intensity (max $\approx 330$ vehicles/hour): 2001, 4010, and 4026.
    \item Group of detectors with Medium traffic intensity (max $\approx 600$ vehicles/hour): 2005 and 4043.
    \item Group of traffic detectors with High traffic intensity (max $\approx 1100$ vehicles/hour):  4063, 8009, 8008, 1032, 1033.
    \end{itemize}

\begin{figure}[ht] 
    \centering
    \includegraphics[width=0.45\textwidth]{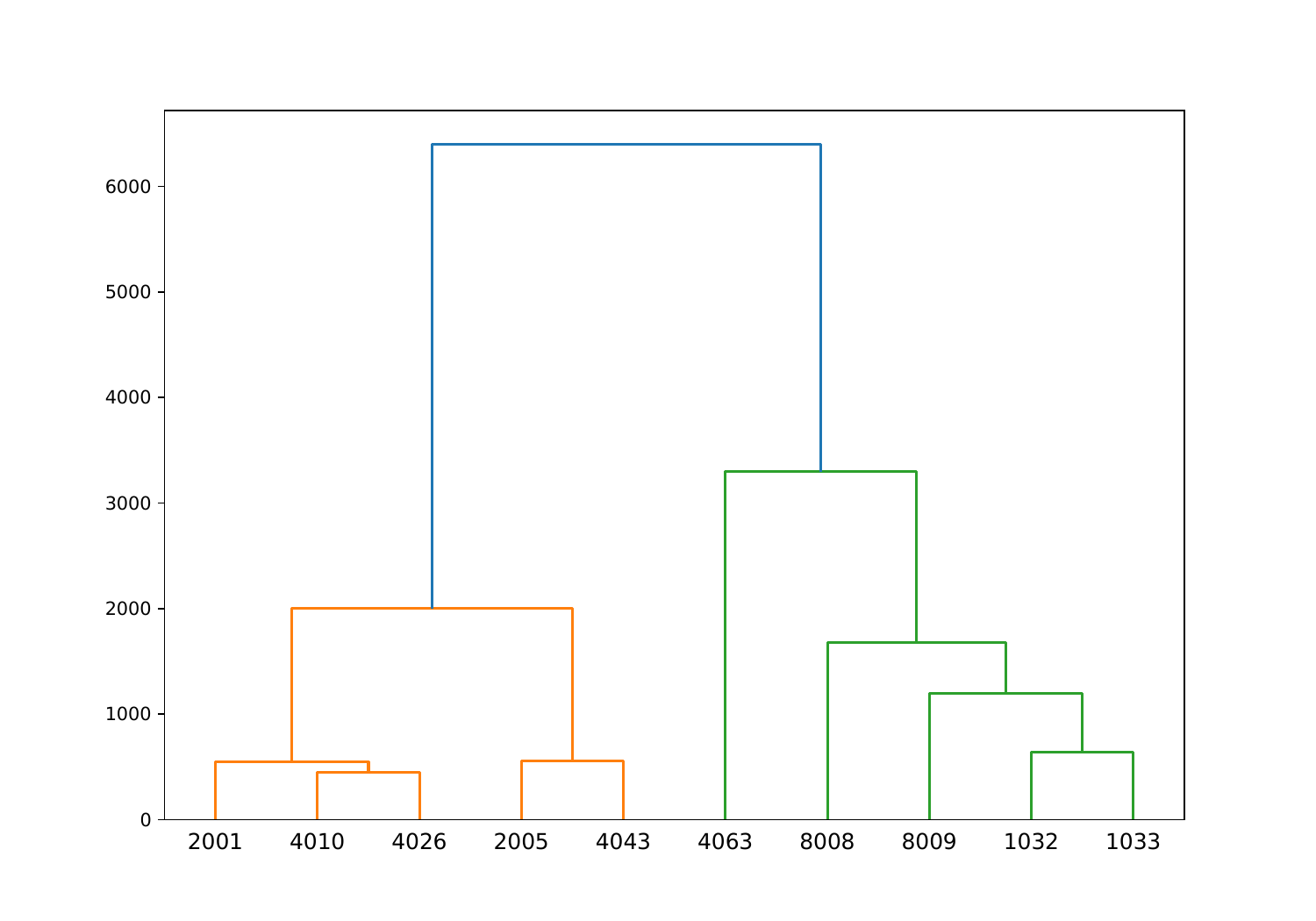} 
    \caption{Dendrogram illustrating clustering based on traffic volume similarity. Y-axis: Distance (vehicles/hour). X-axis: Detector ID.}
    \label{fig:dendro_volume}
\end{figure}   

    \item \textbf{Affinity-Based Neighbors (Pattern Similarity):} In this strategy, nodes are grouped according to the shape of their 24-hour traffic profiles. Similarity is computed by first standardizing each detector's 24-hour traffic profile (to have zero mean and unit variance) and then applying hierarchical clustering using Ward's linkage method with Euclidean distance on these standardized profiles. This method, visualized in the dendrogram in Fig.~\ref{fig:dendro_pattern}, aims to group regions with similar temporal traffic dynamics, regardless of the absolute traffic volume. It is worth noting that, for the 10 selected traffic detectors, both strategies yielded the same partition of traffic detectors.

\begin{figure}[ht]
\centering
\includegraphics[width=0.45\textwidth]{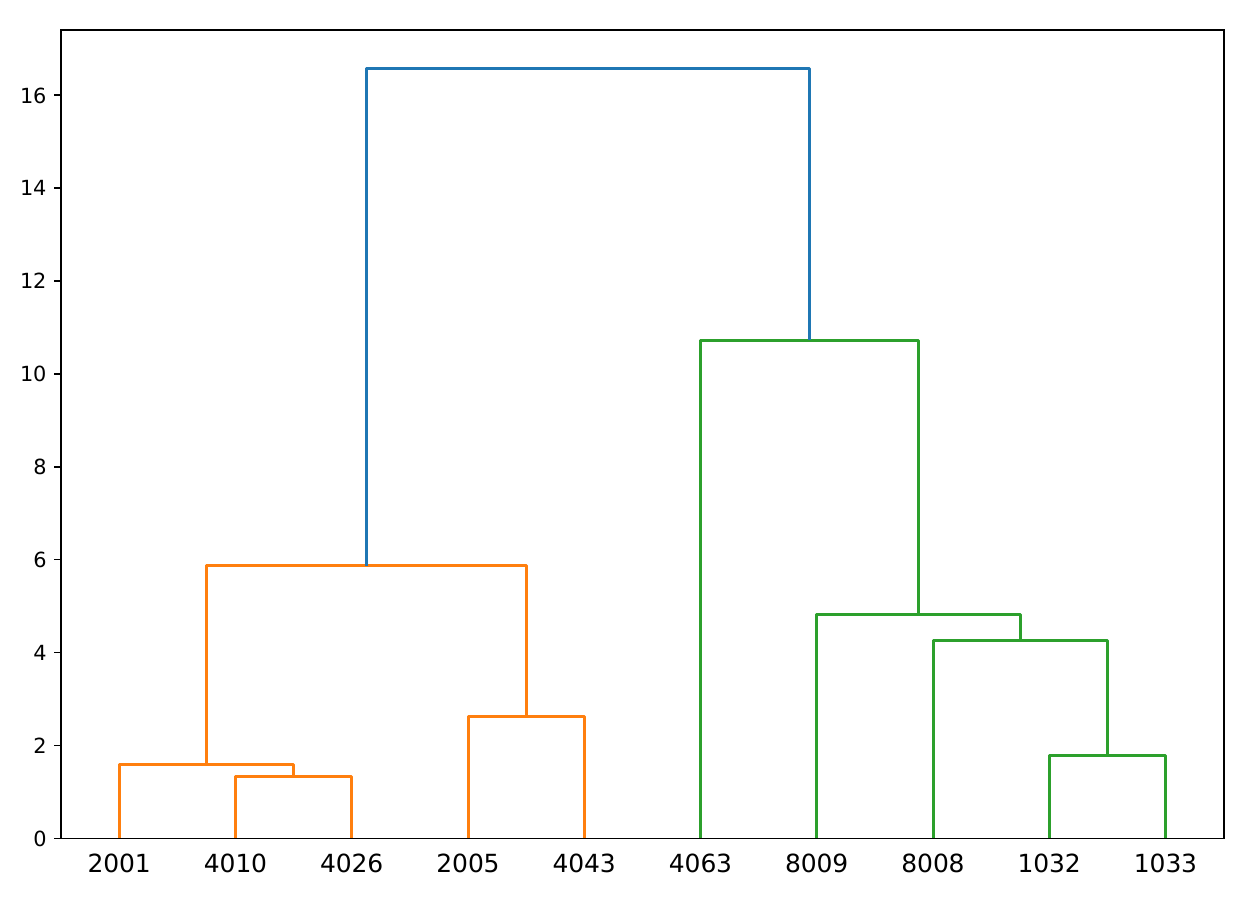} 
\caption{Dendrogram illustrating clustering based on traffic pattern similarity. Y-axis: Standardized distance. X-axis: Detector ID.}
\label{fig:dendro_pattern}
\end{figure}

\end{itemize}

\subsection{Evaluation Metrics}
\label{sec:metrics}

To evaluate the accuracy of the generated traffic profiles, we conduct a quantitative performance assessment of our traffic generation system using three primary error metrics, each tailored to a specific validation context.

\textbf{Relative Error (RE)}. It is used during training and testing of traffic generation over one-hour intervals. For a given detector, this metric quantifies the deviation between the target vehicle count and the corresponding output from the SUMO simulation. Formally, it is defined as:
\begin{equation}
\label{eq:rel_error}
\text{RE} = T - D,
\end{equation}
where \( T \) denotes the target vehicle count for a specific hour, and \( D \) is the output obtained from the detector after simulation. This signed error captures both underestimation and overestimation, and is also used to track policy improvement over training rounds in the decentralized learning setup. This metric captures whether the system overestimates or underestimates the required traffic. It is also used to evaluate the evolution of performance in model exchange rounds in the decentralized learning setup (see Algorithm \ref{alg:algo1h}).

\textbf{Average Relative Error (ARE)}. This metric across multiple one-hour test episodes provides an aggregate measure of the performance of Algorithm \ref{alg:algo1h} beyond its training set. Let \( RE_j \) denote the relative error in the \( j \)-th test. The ARE is defined as:

\begin{equation}
\label{eq:are}
\text{ARE} = \frac{1}{N} \sum_{j=1}^{N} \left| RE_j \right| = \frac{1}{N} \sum_{j=1}^{N} \left| T_j - D_j \right|,
\end{equation}

\noindent where \( N \) is the total number of test episodes; \( T_j \) and \( D_j \) are the target and output vehicle counts, respectively, for the \( j \)-th test. This metric summarizes the average deviation magnitude across a set of tests and offers a robust measure of generalization and reliability for Algorithm \ref{alg:algo1h}.

\textbf{Mean Absolute Error (MAE)}. It is used for extended validation of full-day (24-hour) traffic generation tasks, as carried out by Algorithm \ref{alg:algo24h}. It quantifies the average magnitude of hourly deviations across an entire daily traffic profile. The MAE is formally defined as:

\begin{equation}
\label{eq:mae}
\text{MAE} = \frac{1}{24} \sum_{i=0}^{23} \left| T_i - D_i \right|,
\end{equation}

\noindent where \( T_i \) and \( D_i \) are the target and output vehicle counts, respectively, for hour \( i \in \{0,\dots,23\} \). This metric provides a robust indicator of how accurately the system replicates full-day traffic patterns.

To further assess the consistency of Algorithm \ref{alg:algo24h}, multiple 24-hour simulations can be conducted. The resulting set of MAE values can then be analyzed statistically (e.g., computing their mean and standard deviation) to evaluate the stability and generalization capability of the trained model across varying target profiles.
   
\section{RESULTS AND DISCUSSION}
\label{sec:results}

To comprehensively assess the performance and robustness of the proposed framework for generating realistic 24-hour traffic profiles, this section presents a structured analysis of experimental results across multiple configurations. We begin by validating the core algorithms in isolated scenarios to evaluate their ability to replicate real traffic patterns. We then explore different decentralized training strategies and assess their relative impact on system-wide learning quality. Subsequently, we analyze the behavior of the fully decentralized framework under diverse traffic conditions and detector profiles. Lastly, we compare the proposed methodology against existing baseline solutions to highlight its advantages in terms of accuracy and generalization.
    
\subsection{SUMO Validation of Single-Zone Traffic Generation}

To validate the effectiveness of the proposed traffic generation framework using reinforcement learning, we conducted an experiment focusing on a single detector, chosen among the group of detectors with the highest traffic intensities: \textbf{Detector 1032}. This detector, along with its associated area of influence, defined by its Voronoi-based TAZ, was selected as the testbed for evaluating both Algorithm~\ref{alg:algo1h} (hour-wise traffic generation) and Algorithm~\ref{alg:algo24h} (extended 24-hour traffic generation).

In this setup, the RL agent was trained for 500 episodes to learn the historical traffic intensity targets specific to Detector 1032 over a typical 24-hour weekday cycle. Operating within its assigned region, the agent adjusted the number of vehicles injected per hour to minimize the discrepancy between simulated and target traffic volumes.

\begin{figure}[t] 
\centering 
\includegraphics[width=0.8\linewidth]{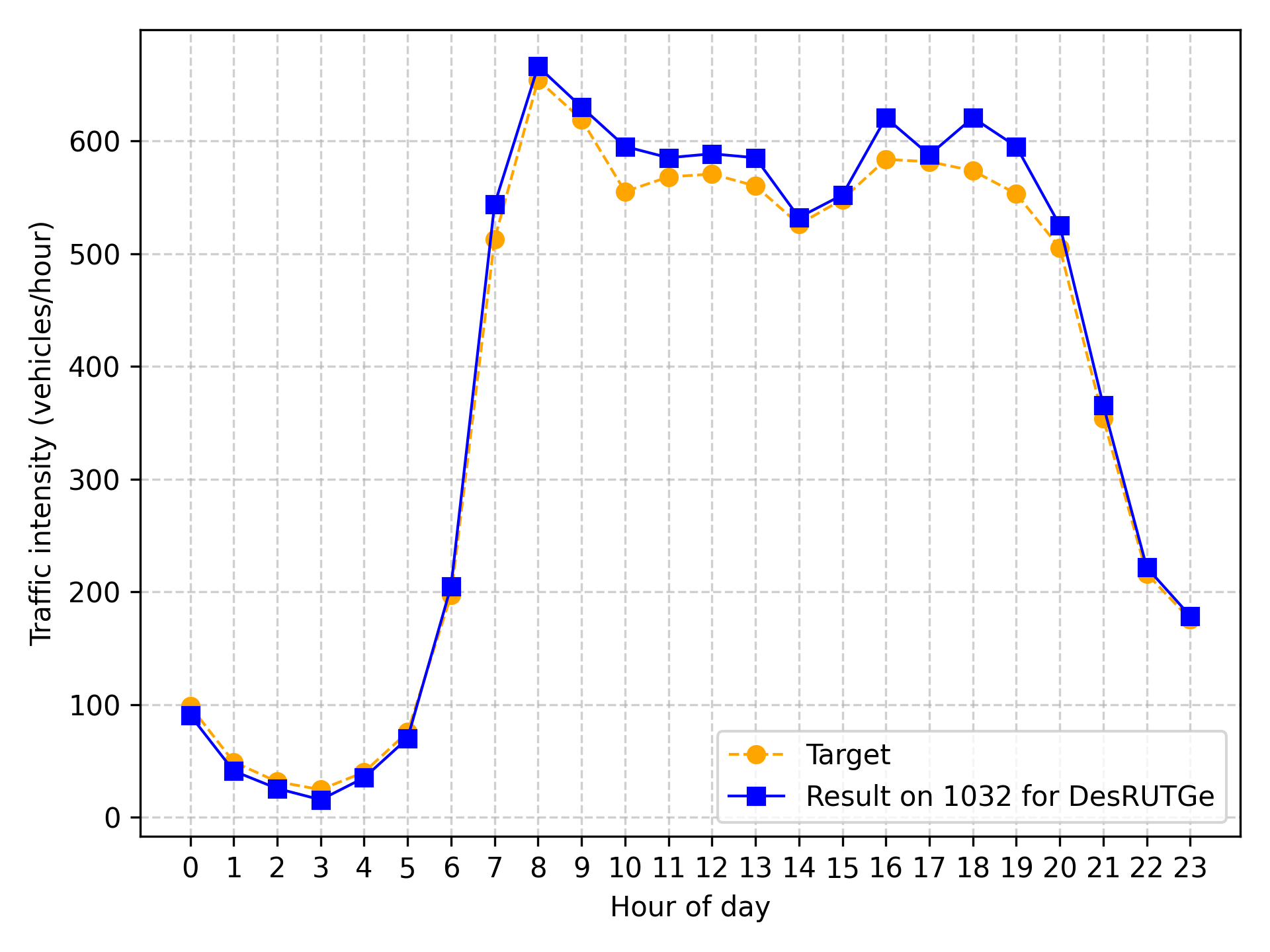} \caption{Validation of traffic generation for Traffic Detector 1032 over a 24-hour period. The plot compares the target traffic intensities (derived from historical data) with the values generated by the RL agent using Algorithms~\ref{alg:algo1h} and~\ref{alg:algo24h}. The close alignment between both curves demonstrates the agent’s capability to reproduce realistic traffic patterns within the region associated with the detector.}
\label{fig:validation_detector_1032} 
\end{figure}

Figure~\ref{fig:validation_detector_1032} illustrates the results of this validation experiment. The curves show the alignment between the generated traffic and the real-world target intensities. The close correspondence observed confirms that the agent successfully learns to reproduce realistic traffic dynamics within the selected zone.

\begin{figure*}[!t]
    \centering
    \subfloat[Volume Similarity]{\includegraphics[width=0.33\textwidth]{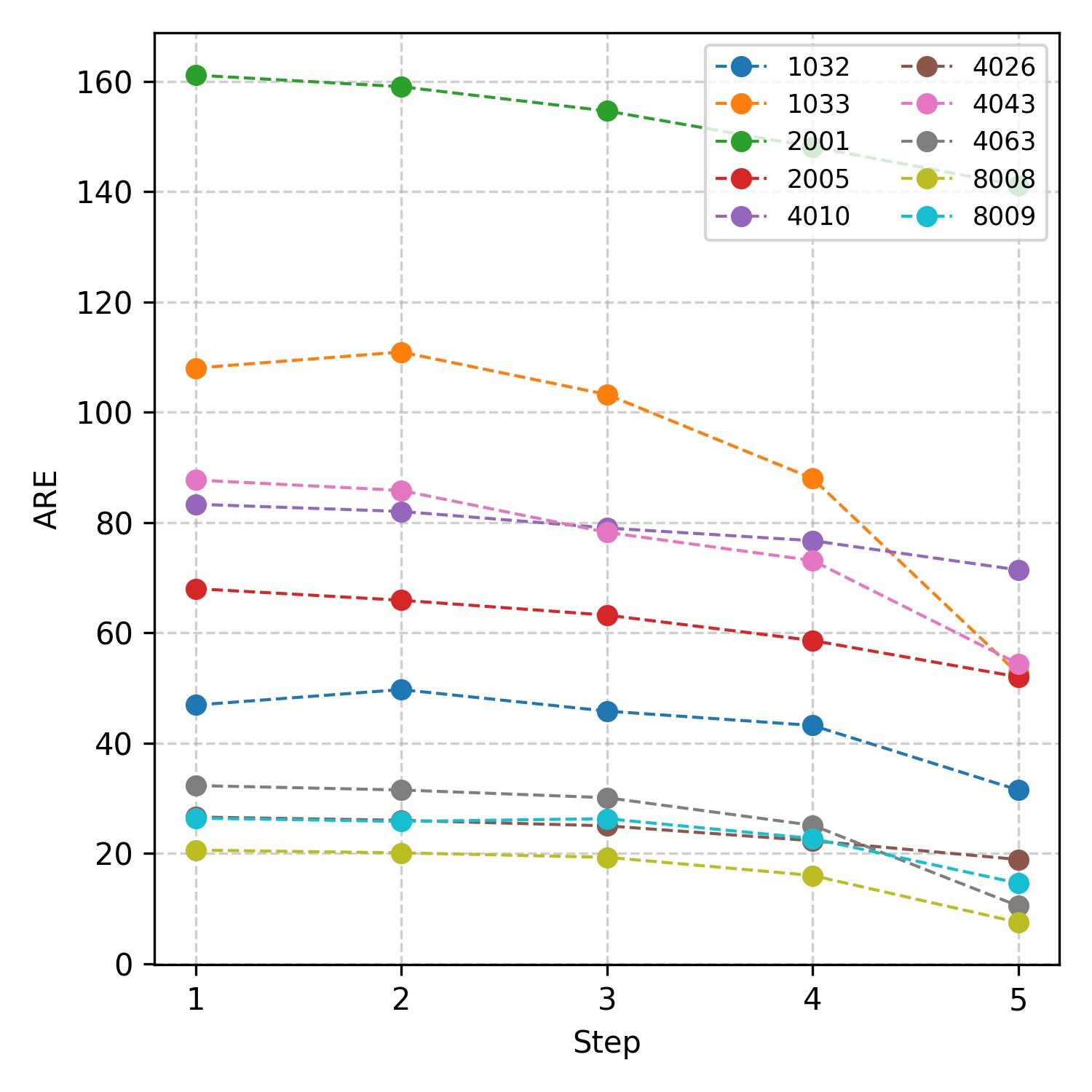}\label{fig:3subfig1}}
    \hfil
    \subfloat[Pattern Similarity]{\includegraphics[width=0.33\textwidth]{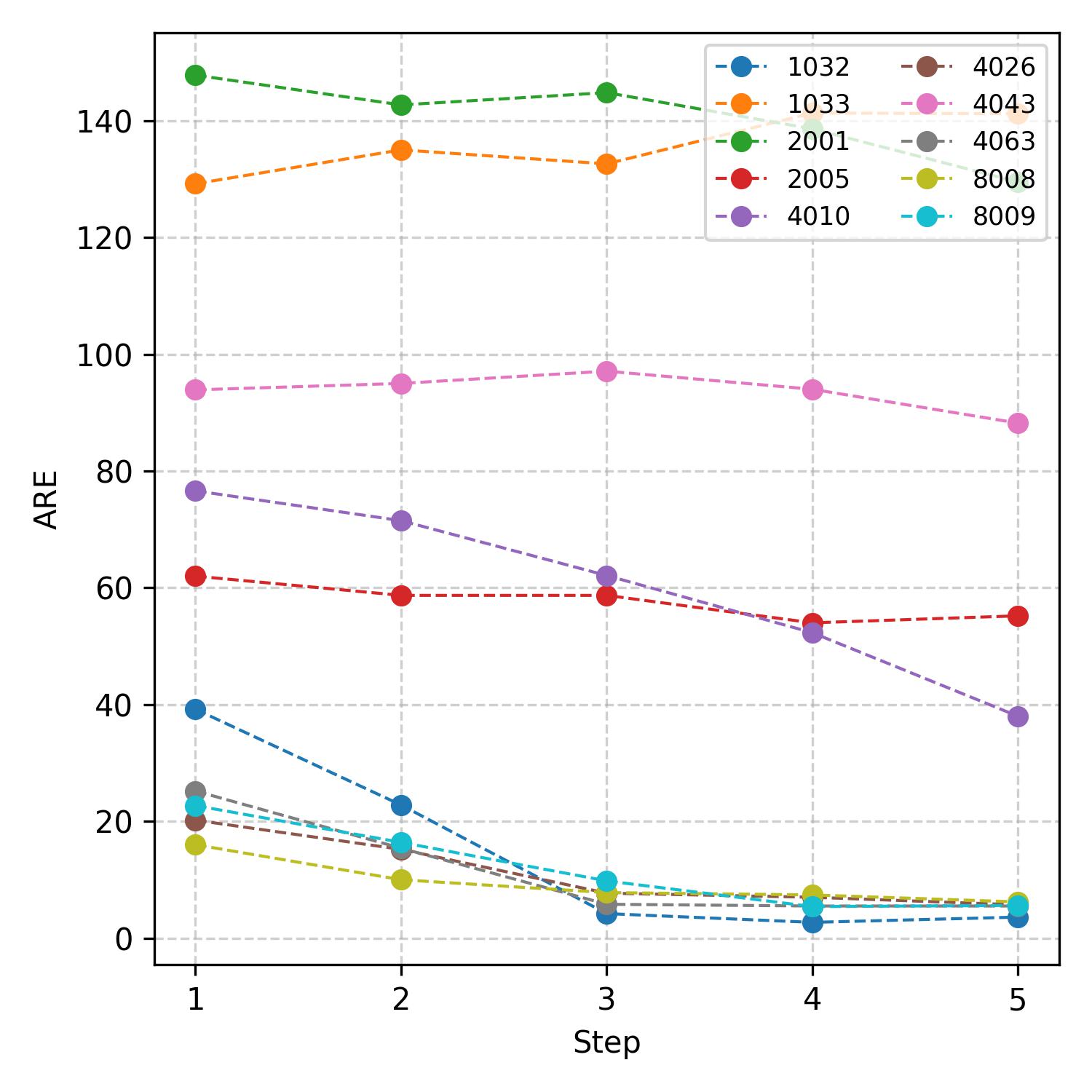}\label{fig:3subfig2}}
    \hfil
    \subfloat [Geographic Neighbors]{\includegraphics[width=0.33\textwidth]{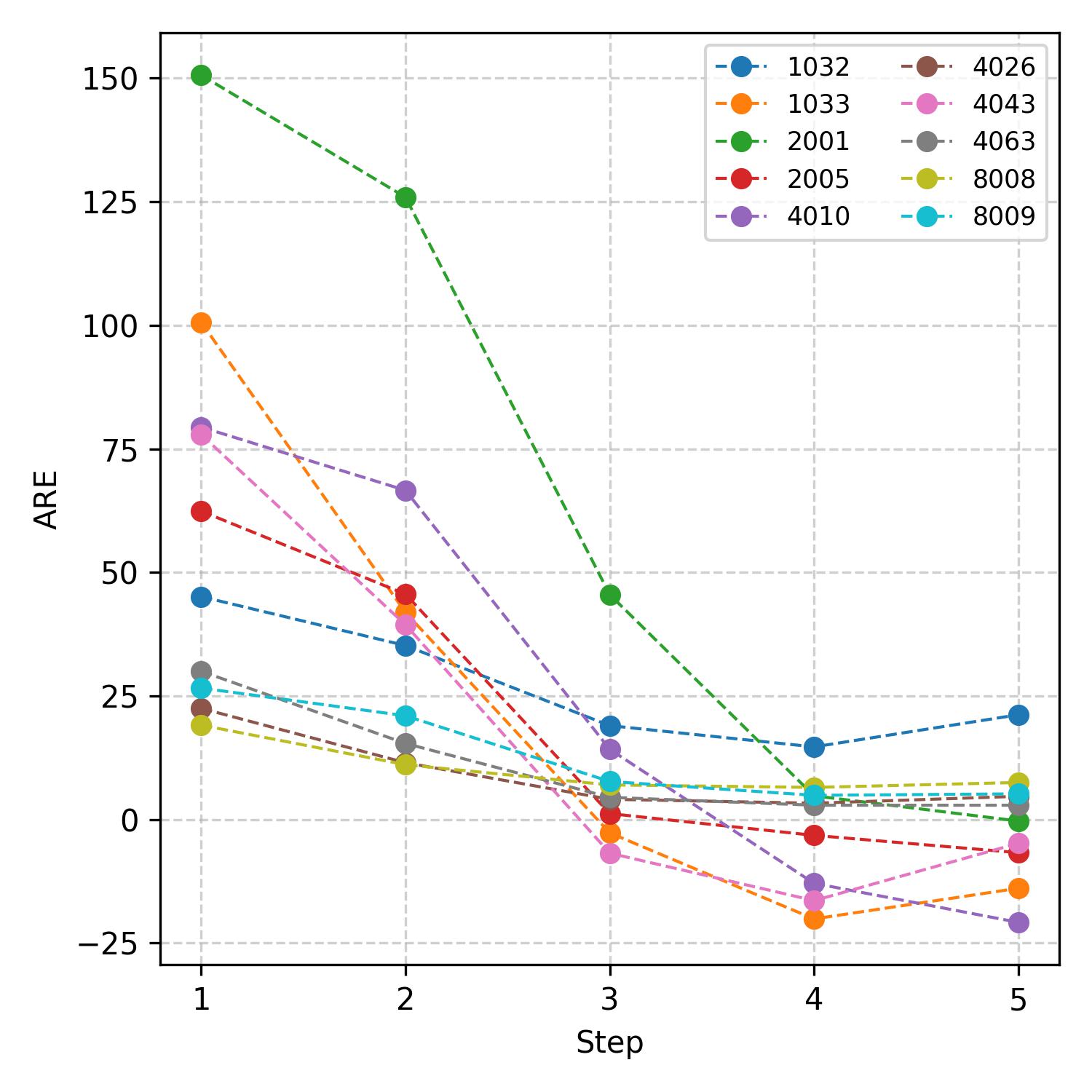}\label{fig:3subfig3}}
    \caption{Comparison of \textbf{ARE} across training rounds for three model-sharing strategies: (a) Affinity-Based Neighbors (Volume Similarity), where exchange occurs among nodes with similar overall traffic volumes; (b) Affinity-Based Neighbors (Pattern Similarity), where nodes share models based on similarity in their 24-hour traffic patterns and (c) Geographic Neighbors, where nodes exchange models with adjacent TAZ regions in the Voronoi diagram. The results demonstrate that the Geographic Strategy achieves better convergence and lower errors over time, in all the traffic detectors.}
    \label{fig:exchange_results}
\end{figure*}

This targeted validation demonstrates that both Algorithms~\ref{alg:algo1h} and~\ref{alg:algo24h} effectively generate traffic patterns closely adhering to real-world data when applied to a single detector and its corresponding region. These results establish a solid foundation for scaling the approach to more complex, heterogeneous multi-zone scenarios.

\subsection{Model exchange validation}
To assess the effectiveness of  the three model-sharing strategies introduced in Section \ref{sec:experimental setup}-\ref{subsec:decentralized}, each strategy is tested over 5 training rounds with 100 episodes of training. After each round, the PPO agent for each node is evaluated on a test set of 100 traffic intensity targets. These targets are linearly sampled between the minimum and maximum historical values for the node. For each test case, the agent attempts to generate traffic that matches the target, and the absolute error is measured.

The evaluation metric used is the \textbf{ARE} computed over the 100 target traffic profiles after each communication round. A decreasing ARE across rounds indicates improved learning performance, driven by effective inter-agent model exchange. Figure~\ref{fig:exchange_results} presents the results, comparing the three exchange strategies.

Experimental results indicate that the \textbf{Geographic Neighbors} strategy consistently outperforms the affinity-based alternatives (Volume and Pattern Similarities). It achieves a more significant reduction in error across rounds, suggesting more effective transfer of policy knowledge among physically adjacent zones. In contrast, both \textbf{Affinity-Based} strategies—though conceptually appealing—exhibit slower convergence, implying that geographic proximity may play a more critical role than traffic similarity in enabling successful collaborative learning within urban traffic environments.

\subsection{DesRUTGe Validation - DFL scenario}

To evaluate the performance of the proposed traffic generation methodology, we applied Algorithm~\ref{alg:algo24h} to each of ten selected traffic detectors distributed across the city of Barcelona. For every detector, the corresponding 24-hour historical traffic intensity profile was used as the target input, enabling a comprehensive assessment of the system's ability to replicate real-world traffic patterns in diverse urban regions.

\begin{figure*}[!t]
    \centering
    \subfloat[Detector 4010 (Low traffic intensity)]{\includegraphics[width=0.33\textwidth]{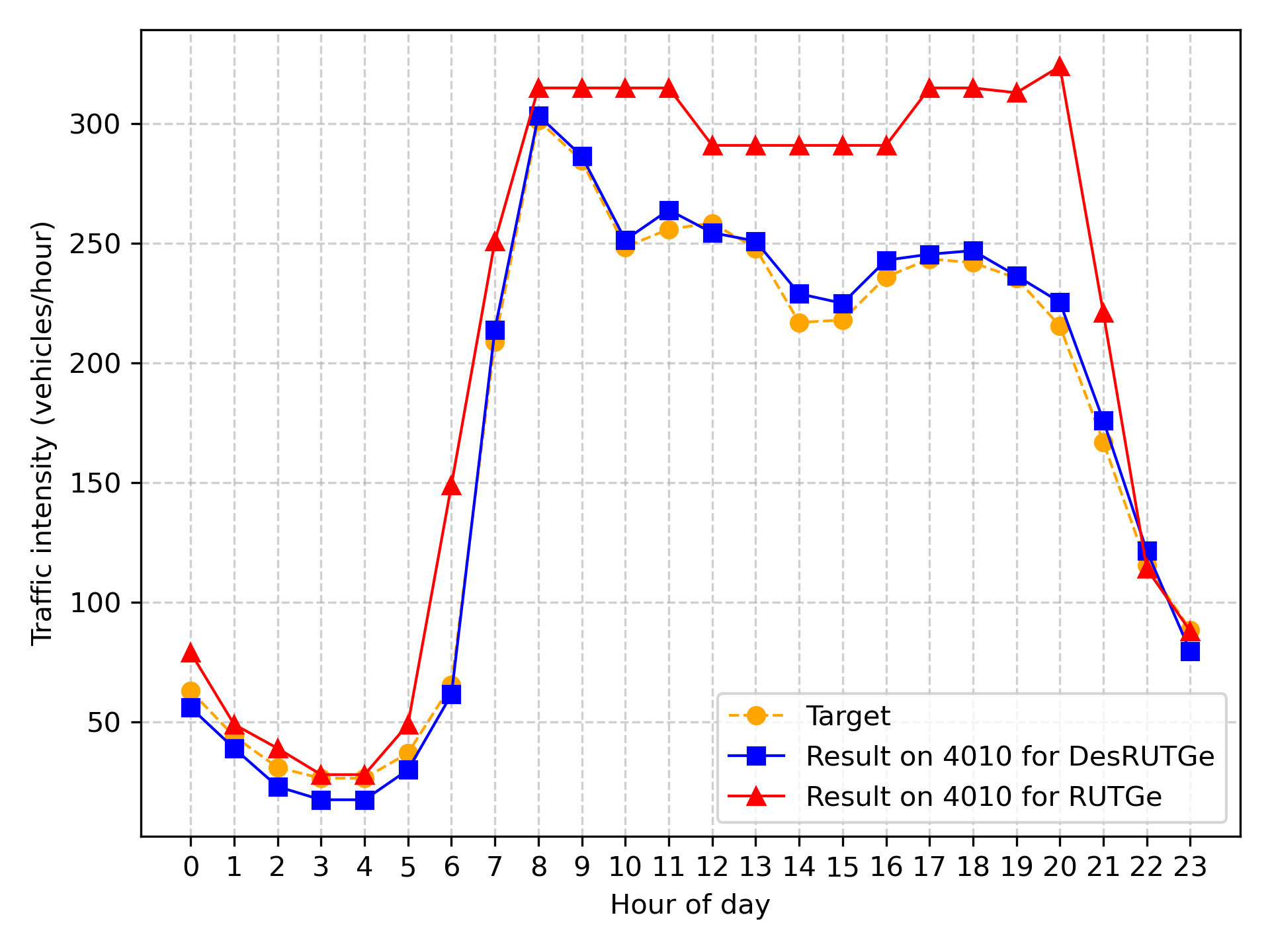}\label{fig:subfig1}}
    \hfil
    \subfloat[Detector 4043 (Medium traffic intensity)]{\includegraphics[width=0.33\textwidth]{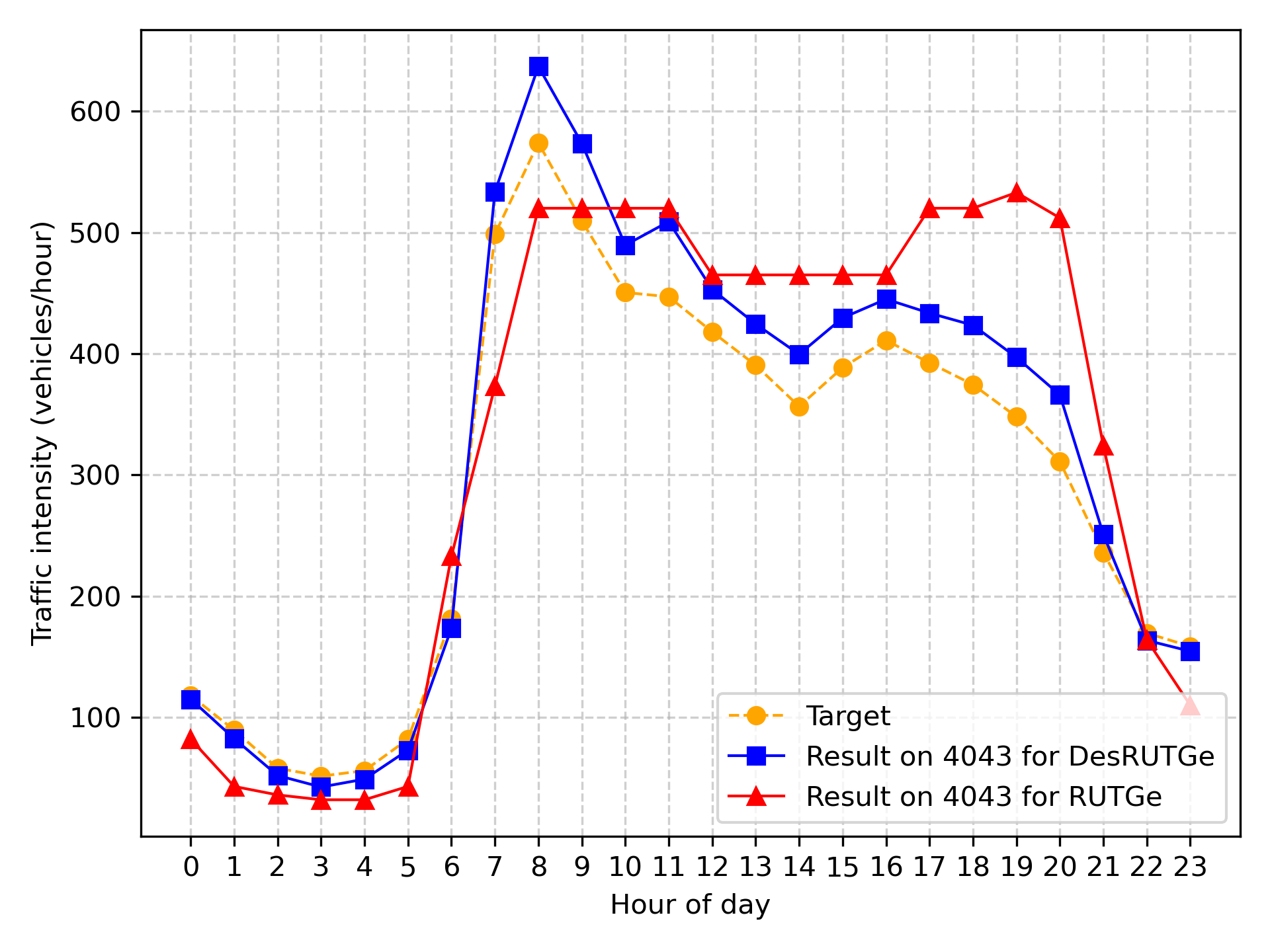}\label{fig:subfig2}}
    \hfil
    \subfloat[Detector 4063 (High traffic intensity)]{\includegraphics[width=0.33\textwidth]{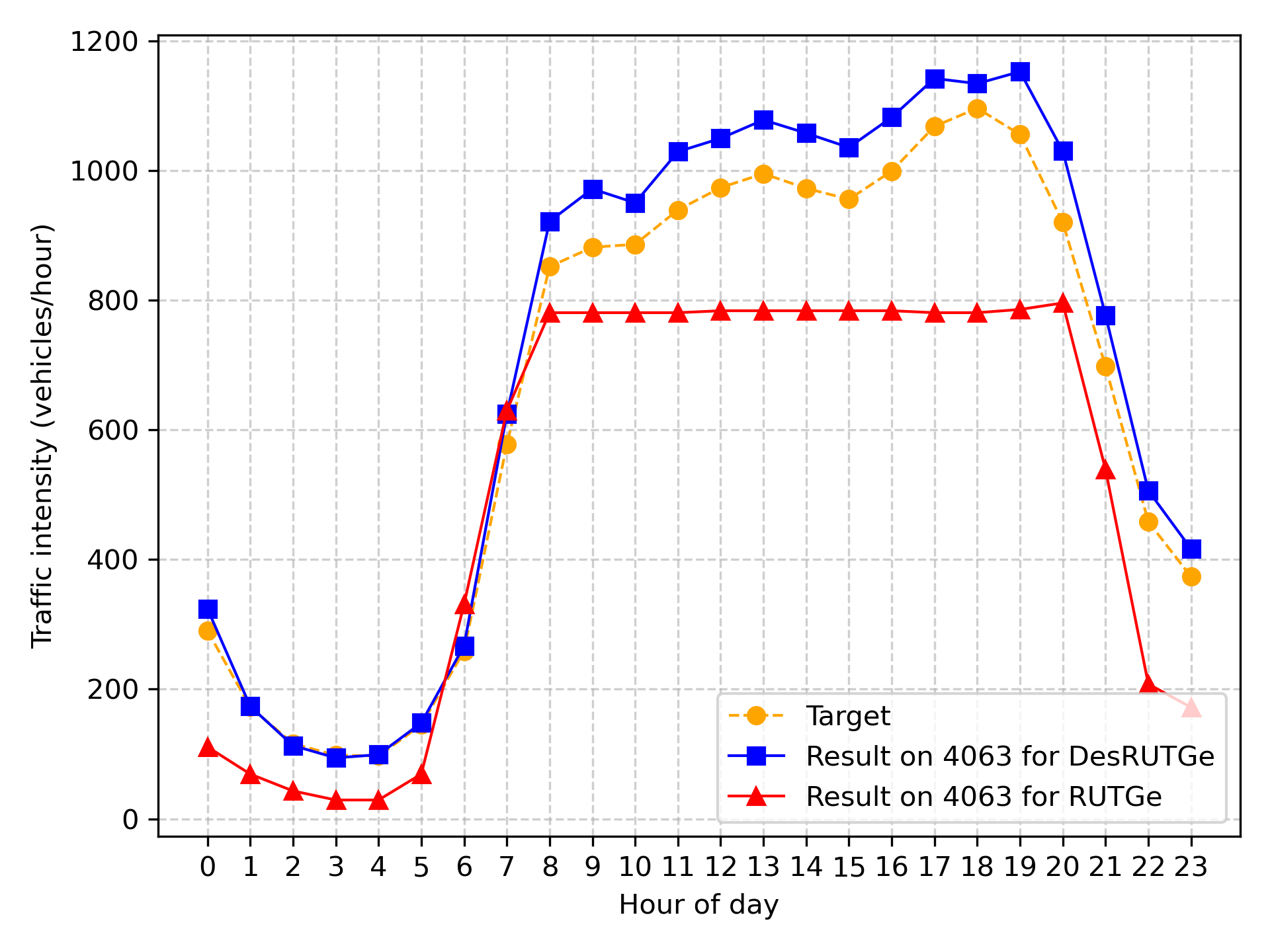}\label{fig:subfig3}}
    \caption{Comparison of generated versus target traffic profiles over a 24-hour period for three representative detectors: 4010 (low-intensity), 4043 (medium-intensity), and 4063 (high-intensity). Each plot includes the output from our proposed DFL-based system DesRUTGe and from the previous centralized method RUTGe. The results show that DesRUTGe approach more accurately follows the target profiles across varying traffic intensities, highlighting the improved adaptability and generalization of Algorithm~\ref{alg:algo24h} over prior centralized solutions.}
    \label{fig:example_outputs}
\end{figure*}

Using the traffic targets as input reference, Algorithm~\ref{alg:algo24h} generated traffic intensity values for each hour, producing a complete daily traffic profile per detector. Following the generation process, we computed the \textit{RE} for each hour by comparing the generated intensity to the target value. The 24 hourly errors were obtained to illustrate the evolution of the total relative error per detector throughout the day, see Fig. \ref{fig:example_outputs}. 

Figure~\ref{fig:example_outputs} illustrates the generated and target traffic curves for three representative detectors, each corresponding to a scenario of low, medium, and high traffic intensities. Detector 4010 was selected to represent the low-intensity traffic scenario, detector 4043 for medium-intensity traffic, and detector 4063 for high-intensity traffic. These plots demonstrate that the generated traffic profile with our proposed DFL-based system DesRUTGe (blue lines) accurately approximates the target traffic profiles (yelow lines), achieving low cumulative deviations in all cases.

\subsection{Comparison with other state-of-the-art works}
To assess the effectiveness of our traffic generation framework named DesRUTGe, we compare its performance against a standard baseline: the \texttt{routeSampler.py} utility provided by SUMO~\cite{SUMO}. This tool is commonly used to generate vehicle insertion flows that approximate given traffic count data by sampling from predefined origin-destination routes and scaling them to match observed counts. While useful for simple calibration tasks, \texttt{routeSampler.py} lacks adaptive learning mechanisms and does not optimize vehicle distributions dynamically over time.

In our experiment, we applied \texttt{routeSampler.py} under the same 24-hour target conditions defined for each detector, using identical traffic demand profiles. The \textbf{MAE} between the target and generated traffic was computed following the same methodology described in Section~\ref{sec:metrics}.

For a fair comparison, the reinforcement learning agent was trained over five rounds, each comprising 100 training episodes. This setup ensured parity in simulation effort and exposure between both approaches.

Table~\ref{tab:relative_error_comparison} presents the results obtained for both our proposed tool and SUMO's \texttt{routeSampler.py} after conducting 10 independent executions per method, using the same 24-hour target profiles for each of the 10 detectors. For each detector (row), the table reports the mean and standard deviation of the resulting \textbf{MAE} values across the executions. The results clearly demonstrate the superior performance of our approach, which consistently achieves lower deviations from the target traffic patterns. On average, our method yields a significantly lower \textbf{MAE} of 25.88 compared to 69.32 obtained by \texttt{routeSampler.py}, indicating a more accurate and reliable traffic generation process.

\begin{table*}[htp]
\centering
\caption{Comparison of mean ($\mu$) and standard deviation ($\sigma$) of the \textbf{MAE} over 10 executions of the proposed traffic generation method DesRUTGe, compared to the SUMO's \texttt{routeSampler.py}, evaluated on 24-hour traffic targets for each of the 10 detectors. Detectors are sorted in ascending order of traffic intensity, and grouped accordingly. Results show that our method achieves significantly lower average errors, demonstrating its superior accuracy in generating real traffic patterns.}

\label{tab:relative_error_comparison}
\begin{tabular}{l lcc}
\toprule
\textbf{Traffic Intensity} & \textbf{Detector ID} & \textbf{MAE–DesRUTGe (\(\mu \pm \sigma\))} & \textbf{MAE–routeSampler (\(\mu \pm \sigma\))} \\
\rowcolor[gray]{0.95}
Low & 4010 & \(6.67 \pm 0.69\)   & \(22.97 \pm 8.46\) \\
\rowcolor[gray]{0.95}
Low & 4026 & \(6.58 \pm 0.01\)   & \(10.85 \pm 4.01\) \\
\rowcolor[gray]{0.95}
Low & 2001 & \(6.85 \pm 0.70\)   & \(16.47 \pm 7.09\) \\
\rowcolor[gray]{0.90}
Medium & 2005 & \(33.17 \pm 0.32\)  & \(42.47 \pm 7.40\) \\
\rowcolor[gray]{0.90}
Medium & 4043 & \(29.82 \pm 0.14\)  & \(67.74 \pm 7.18\) \\
\rowcolor[gray]{0.85}
High & 1032 & \(23.59 \pm 3.11\)  & \(101.44 \pm 32.45\) \\
\rowcolor[gray]{0.85}
High & 8009 & \(45.96 \pm 0.01\)  & \(154.10 \pm 20.72\) \\
\rowcolor[gray]{0.85}
High & 1033 & \(19.66 \pm 2.25\)  & \(96.82 \pm 43.07\) \\
\rowcolor[gray]{0.85}
High & 8008 & \(31.83 \pm 2.70\)  & \(146.90 \pm 23.28\) \\
\rowcolor[gray]{0.85}
High & 4063 & \(54.71 \pm 0.01\)  & \(33.49 \pm 8.83\) \\
\midrule
\textbf{–} & \textbf{Mean} & \textbf{\(25.88 \pm 0.99\)} & \textbf{\(69.32 \pm 16.25\)} \\
\bottomrule
\end{tabular}
\end{table*}

In the results shown in Fig.~\ref{fig:example_outputs} it is also evident that the centralized method RUTGe fails to accurately replicate the expected traffic patterns for each individual detector. This limitation stems from its design, which relies on computing a global average of all detector targets to define a single unified traffic target per hour across the entire simulation. As a result, while RUTGe may yield acceptable average performance at the city-wide level, it lacks the granularity required to match localized traffic demands. The generated outputs tend to converge toward the same average value across all detectors, disregarding their specific traffic characteristics. This behavior highlights the architectural limitation of RUTGe in handling spatial traffic heterogeneity, and underscores the superior precision of our proposed DesRUTGe framework, which is explicitly designed to align traffic generation with the target of each individual zone.

The decentralized approach of our DesRUTGe proposal has been shown through previous results to enable a more scalable and robust collaborative learning environment that is less susceptible to single-point failures. This approach also allows for the generation of more accurate synthesized traffic in each city area.  However, adopting this methodology has other disadvantages that need to be addressed. First, decentralizing the federation process requires sharing more intermediate models between detectors in each round. This increases communication overhead, impacts the digital carbon footprint, and increases energy consumption. To this end, we will consider adopting rapid information dissemination techniques using protocols such as Gossip in future work.

Under the proposed DFL-based DesRUTGe traffic generation framework, the fully decentralized training process—conducted over 10 nodes and 5 rounds, each consisting of 100 training episodes—required a total runtime of 3.34 hours. This process was executed on a Dell PowerEdge R7525 server, whose specifications are detailed in Section~\ref{sec:experimental setup}-\ref{sec:implementation}.

In contrast, the former centralized RUTGe version~\cite{guillenrutge} completed an equivalent training task in 1.4 hours on the same machine (see red lines in Fig.~\ref{fig:example_outputs}). This notable increase in execution time is not primarily due to the decentralized nature of the framework, but rather to the training traffic intensity values used. While the centralized RUTGe approach employs a single average traffic intensity across all detectors, the DesRUTGe framework allows each detector to train using its individual traffic intensity profile. These localized profiles are often higher than the average, leading to more demanding simulations and, consequently, longer runtimes.

Nonetheless, the increase in computational cost is justified by the enhanced accuracy achieved by DesRUTGe, as illustrated in Fig.~\ref{fig:example_outputs}. These findings highlight the trade-off between computational efficiency, scalability, and modeling precision enabled by the proposed decentralized architecture.

\section{CONCLUSIONS AND FUTURE WORK}
\label{sec:conclusions}
In this work, we propose a novel decentralized traffic generation framework based on Deep Reinforcement Learning (DRL), aimed at synthesizing realistic urban traffic patterns by leveraging historical data from multiple traffic detectors. The goal is to generate traffic profiles suitable for use in simulation environments, enabling researchers to evaluate their proposals under realistic, data-driven conditions. The proposed decentralized framework demonstrates strong performance in reproducing accurate traffic patterns and offers a suitable approach for modeling complex urban mobility dynamics.

The proposed approach partitions the simulation map into non-overlapping zones using a Voronoi-based decomposition centered on each traffic detector, enabling localized policy learning and traffic generation with higher spatial fidelity. By deploying local PPO agents associated with each detector and its corresponding Traffic Analysis Zone (TAZ), we demonstrated the ability to generate realistic hourly traffic intensities over a complete 24-hour period. Algorithm~\ref{alg:algo24h} consistently outperforms the traditional SUMO \texttt{routeSampler.py} tool, achieving significantly lower relative errors. In our experiments using real-world traffic data provided by the City Hall of Barcelona, the proposed method yielded error reductions nearly three times lower on average, confirming its effectiveness for fine-grained and data-driven traffic synthesis.

Furthermore, we investigated the advantages of decentralized learning by evaluating different inter-node model exchange strategies. Experimental results revealed that geographically-based neighbor sharing consistently outperforms affinity-based alternatives in terms of both convergence speed and error reduction. These findings suggest that spatial proximity between zones facilitates more effective knowledge transfer during training, enhancing the overall performance of the decentralized framework.

Our traffic generation system not only enhances the realism of simulated urban mobility scenarios, but also lays the groundwork for scalable, region-aware modeling applicable to smart city planning and adaptive traffic control. Future directions include exploring dynamic TAZ definitions, designing adaptive reward functions tailored to each zone, and integrating real-time data streams to enable online learning in live simulation environments.

This research lies at the intersection of Deep Reinforcement Learning (DRL), Decentralized Federated Learning (DFL), and urban traffic simulation, addressing the critical need for realistic, scalable, and privacy-preserving traffic modeling tools. These capabilities are of global importance, as cities around the world strive to optimize transportation systems for both efficiency and sustainability. While this work has focused on the core DFL-DRL mechanism and demonstrated its effectiveness for traffic generation, future research should explore advanced DFL aggregation techniques to improve robustness against non-IID data distributions and potential Byzantine behaviors. Additionally, a deeper analysis of communication efficiency in decentralized settings remains a valuable direction to further enhance the practicality and scalability of the framework.

\section*{Acknowledgment}

This work was supported by the grant PID2020-113795RB-C33 funded by MICIU/AEI/10.13039/501100011033 (COMPROMISE project), the grant PID2023-148716OB-C31 funded by MCIU/AEI/10.13039/501100011033 (DISCOVERY project); and ``TRUFFLES: TRUsted Framework for Federated LEarning Systems, within the strategic cybersecurity projects (INCIBE, Spain), funded by the Recovery, Transformation and Resilience Plan (European Union, Next Generation)''. Additionally, it also has been funded by the Galician Regional Government under project ED431B 2024/41 (GPC). Also, by the project "Anonymization technology for AI-based analytics of mobility data (MOBILYTICS)" TED2021-129782B-I00 funded by MCIN/AEI/10.13039/501100011033 and by the European Union NextGenerationEU/ PRTR; Also, by the project "MultiMO: Datos MultiSectoriales para la Movilidad Obligada" TSI-100123-2024-60 (European Union, NextGenerationEU); Also, by predoctoral scholarship for the training of research personnel associated with the "Generación de Conocimiento" Project PRE2021-099830; Finally, by the Generalitat de Catalunya under AGAUR grant "2021 SGR 01413".


\end{document}